\documentclass[10pt,twocolumn,letterpaper]{article}

\usepackage[pagenumbers]{cvpr} 

\usepackage[dvipsnames]{xcolor}
\usepackage{booktabs}
\usepackage{adjustbox}
\usepackage{csquotes}

\newcommand{\datasetname}{SynMirror\xspace}
\newcommand{\testsetname}{MirrorBench\xspace}
\newcommand{\methodname}{MirrorFusion\xspace}

\usepackage{tocloft}
\usepackage{algorithm} 
\usepackage{algpseudocode} 
\usepackage[section]{placeins}

\definecolor{cvprblue}{rgb}{0.21,0.49,0.74}
\usepackage[pagebackref,breaklinks,colorlinks,citecolor=cvprblue]{hyperref}

\def\paperID{250} 
\def\confName{3DV\xspace}
\def\confYear{2025\xspace}

\title{
Reflecting Reality: Enabling Diffusion Models to Produce\\Faithful Mirror Reflections
}

\author{
    Ankit Dhiman$^{1,2}$\thanks{} \quad Manan Shah$^{1}$\footnotemark[1] \quad Rishubh Parihar$^1$ \quad Yash Bhalgat$^3$ \\ 
    Lokesh R Boregowda$^2$ \quad R Venkatesh Babu$^1$ \\ \\
    $^1$Vision and AI Lab, IISc Bangalore \quad $^2$Samsung R \& D Institute India - Bangalore \quad \\ $^3$Visual Geometry Group, University of Oxford
}

\begin{document}


\twocolumn[{%
		\renewcommand\twocolumn[1][]{#1}%
		\maketitle
            \vspace{-3em}
		\begin{center}
			\includegraphics[width=0.98\textwidth]{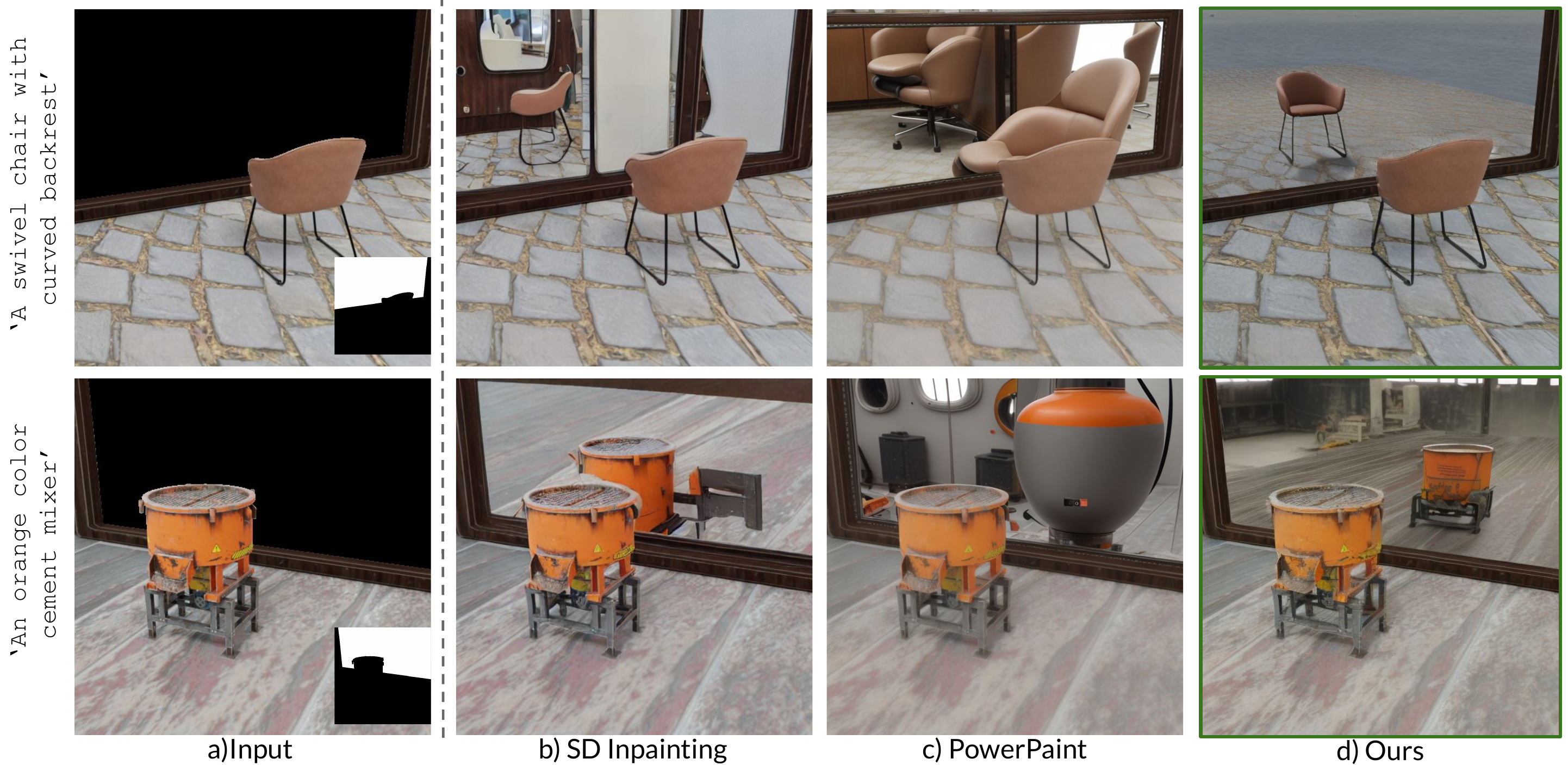}
\vspace{-0.5em}
   \captionsetup{type=figure}
			\captionof{figure}{ We present~\methodname, a diffusion-based inpainting model, which generates high-quality geometrically consistent and photo-realistic mirror reflections given an input image and a mask depicting the mirror region. Our method shows superior quality generations as compared to previous state-of-the-art diffusion-based text-to-image and inpainting methods. All the images were generated by prefixing the mirror text prompt: \enquote{A perfect plain mirror reflection of } to the input object description.
			}
			\label{fig:teaser}
		\end{center}
            \vspace{-1mm}
	}]
 \let\thefootnote\relax\footnotetext{*Equal Contribution.}
\begin{abstract}
\vspace{-2mm} 
We tackle the problem of generating highly realistic and plausible mirror reflections using diffusion-based generative models. We formulate this problem as an image inpainting task, allowing for more user control over the placement of mirrors during the generation process. To enable this, we create~\datasetname{}, a large-scale dataset of diverse synthetic scenes with objects placed in front of mirrors.~\datasetname{} contains around $198K$ samples rendered from $66K$ unique 3D objects, along with their associated depth maps, normal maps and instance-wise segmentation masks, to capture relevant geometric properties of the scene. Using this dataset, we propose a novel depth-conditioned inpainting method called \emph{\methodname}, which generates high-quality geometrically consistent and photo-realistic mirror reflections given an input image and a mask depicting the mirror region. \emph{\methodname} outperforms state-of-the-art methods on~\datasetname{}, as demonstrated by extensive quantitative and qualitative analysis. To the best of our knowledge, we are the first to successfully tackle the challenging problem of generating controlled and faithful mirror reflections of an object in a scene using diffusion based models.~\datasetname{} and \emph{\methodname} open up new avenues for image editing and augmented reality applications for practitioners and researchers alike. The project page is available at:~\href{https://val.cds.iisc.ac.in/reflecting-reality.github.io/}{https://val.cds.iisc.ac.in/reflecting-reality.github.io/}. 

\end{abstract}
\addtocontents{toc}{\protect\setcounter{tocdepth}{-2}}
\section{Introduction}
\label{sec:intro}
Recent diffusion-based generative models~\cite{podell2023sdxl,sohl2015deep,ho2020denoising, saharia2022palette,ho2022video} have achieved remarkable results, producing visually appealing images across various domains. These models can be conditioned using several modalities, such as text~\cite{ho2020denoising}, depth-image~\cite{tyszkiewicz2023gecco}, sketch~\cite{koley2024s}, for controlled generation~\cite{zhang2023adding,ye2023ip-adapter,zhao2024uni}, enabling various interesting applications. Despite their success, these models struggle to capture subtle geometric cues such as shadows, lighting and specular reflections, as noted in previous studies~\cite{sarkar2023shadows,winter2024objectdrop}. Specifically, the task of generating realistic and controllable mirror reflections remains an unsolved challenge. Existing methods, which tackle perspective issues~\cite{upadhyay2023enhancing, zhao2024uni} and address specular reflections for object removal~\cite{winter2024objectdrop} do not address mirror reflections in particular. 

To illustrate this limitation, we prompt Stable Diffusion-2.1~\cite{rombach2022high} with the instruction to generate a scene with a mirror reflection. Fig.~\ref{fig:mirror-reflect-sd} shows that Stable Diffusion-2.1 fails to generate plausible and consistent mirror reflections. Further, various state-of-the-art inpainting methods such as Stable Diffusion Inpainting~\cite{rombach2022high} and PowerPaint~\cite{zhuang2023task} also fail at the task of generating plausible and controlled reflections, as shown in Fig.~\ref{fig:teaser}~(b) \&~(c).  

In this work, we pose the problem of generating mirror reflections as an \emph{Image Inpainting task}. This formulation provides two distinctive advantages: (1) Posing it as an inpainting method aids the reflection generation process to take cues from the input image and (2) allow greater control on the placement of mirrors. 

Existing datasets for tasks such as mirror segmentation, detection or novel-view synthesis as shown in Tab.~\ref{tab:data_characteristic} contain mirrors reflecting generic backgrounds and also lack the scale making them unsuitable for the task of training generative models for generating photo-realistic mirror reflections. Therefore, to address this, we introduce~\datasetname{}, a training dataset and~\testsetname{}, a benchmark dataset designed to train and evaluate the capability of generative models to produce photo-realistic mirror reflections.\datasetname{} contains $198,204$ samples from rendering $66,068$ unique 3D objects sourced from Objaverse~\cite{objaverse} and Amazon Berkeley Objects (ABO)~\cite{collins2022abo}. 

Samples from~\datasetname{} are created by rendering synthetic scenes with objects placed in front of a mirror using the cycles rendering engine from Blender. To further obtain instance segmentation, depth and normal maps, we utilize \texttt{Blenderproc}~\cite{Denninger2023}, a procedural Blender pipeline for photorealistic rendering. The generated scenes have diverse mirrors, floor textures and backgrounds. 

When recent inpainting methods such as BrushNet~\cite{ju2024brushnet} are fine-tuned on~\datasetname{}, we observe that they fail in generating correct geometry and depth of an object in the mirror reflection as shown in Fig.~\ref{fig:cmp_qual}. 
We hypothesize that providing additional cues, such as depth maps, could help to alleviate the issue in generating geometrically consistent reflections on the mirror. To this end, we propose \emph{\methodname}, a depth conditioned inpainting method that generates high-quality controlled and photo-realistic mirror reflections. Our method significantly outperforms state-of-the-art diffusion-based inpainting methods on~\datasetname{}, as evidenced by extensive quantitative and qualitative evaluations. 

We summarize our contributions below:
\begin{itemize}
    \item To the best of our knowledge, we are the first to address and tackle the challenging problem of generating controlled photo-realistic and geometrically consistent mirror reflections of objects using diffusion models by formulating it as an image inpainting task.
    \item For this purpose, we present \emph{\datasetname}, a large-scale synthetic dataset of objects with accurate mirror reflections. We also create~\testsetname{}, a subset of \emph{\datasetname}, for benchmarking the capabilities of generative models in generating photo-realistic mirror reflections of diverse objects.
    \item We further propose \emph{\methodname}, a novel depth conditioned inpainting method, which produces photo-realistic and controlled mirror reflections in the masked region of an input image, when trained on \emph{\datasetname}.

\end{itemize}

\section{Related Work}

\begin{figure}[!t]
    \centering
    \includegraphics[width=\linewidth]{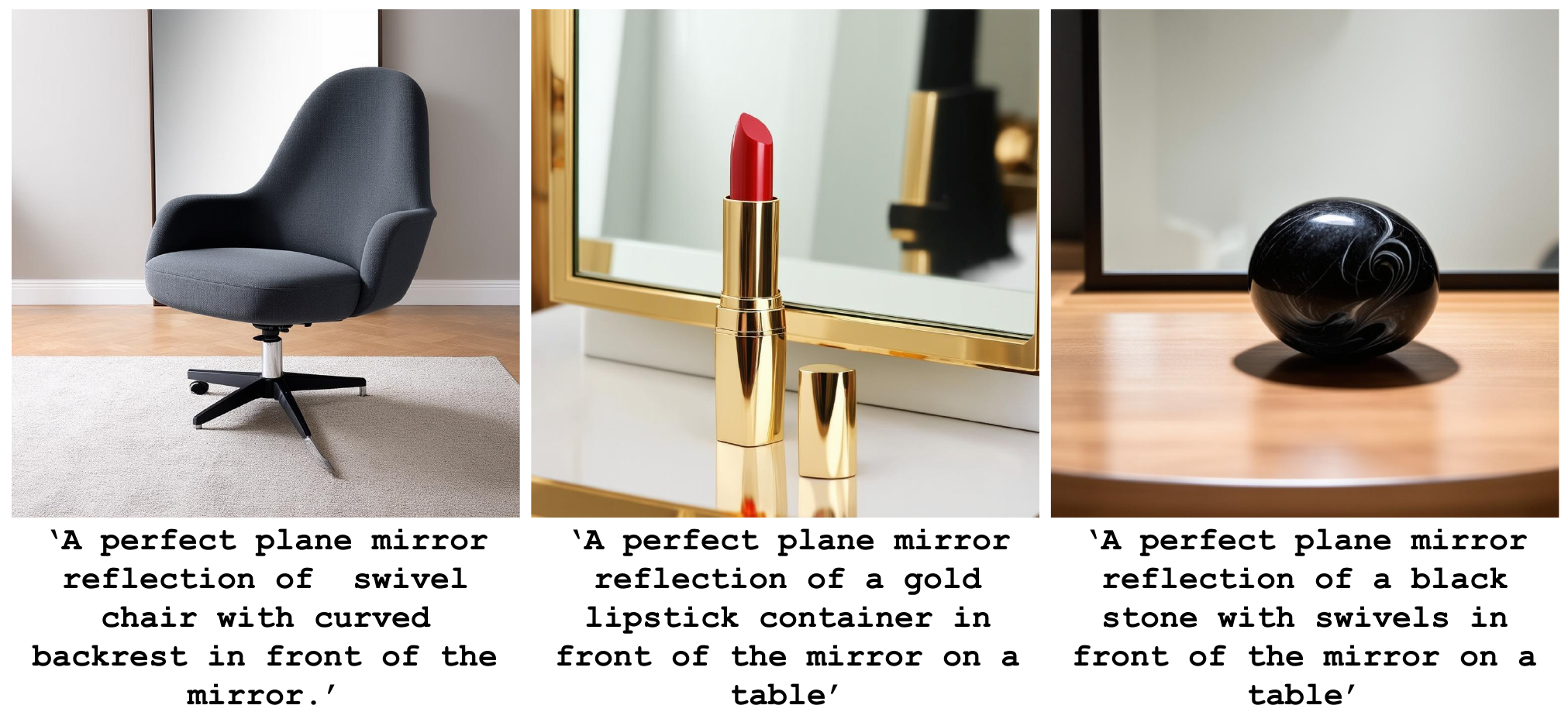}
    \caption{Images generated from Stable Diffusion 2.1~\cite{rombach2022high}. Text-to-image models, when prompted to generate reflections, struggle to generate consistent and controlled mirror reflections.}
    \label{fig:mirror-reflect-sd}
\end{figure}

\noindent
\textbf{Diffusion based generative models.} Diffusion-based models~\cite{sohl2015deep} have revolutionized the field of image synthesis~\cite{ho2020denoising, saharia2022palette}. These generative models are further extended to other modalities such as video~\cite{ho2022video, singer2022make}, audio~\cite{kong2020diffwave}and text~\cite{li2022diffusion}.
Further, text-to-image(T2I) models~\cite{ramesh2022hierarchical,nichol2021glide,rombach2022high,saharia2022photorealistic} have the capability to generate photo-realistic images with any arbitrary text prompt. 
However, these methods do not work for generating realistic reflections, as shown in Fig.~\ref{fig:mirror-reflect-sd}.

\noindent
\textbf{Image inpainting methods.} Recent advancements in diffusion models~\cite{ho2020denoising} have led to significant progress in the inpainting task. Diffusion based inpainting methods~\cite{lugmayr2022repaint,meng2021sdedit,saharia2022palette,avrahami2022blended,corneanu2024latentpaint,zhang2023towards,parihar2024text2place} have shown tremendous improvement in this task compared to GAN based methods~\cite{liu2021pd,zheng2022image}. A common approach to inpaint with diffusion models involves modifying the standard denoising strategy: sampling masked regions from a pre-trained diffusion model and unmasked areas from the given image. While this method produces satisfactory results, it does not generalize to complex scenes and shapes. Stable Diffusion Inpainting~\cite{rombach2022high} fine-tunes a diffusion model by taking the noisy latents, mask and masked image as inputs to the U-Net architecture. Methods like HD-Painter~\cite{manukyan2023hd} further enhance this method. A recent method, BrushNet~\cite{ju2024brushnet}, divides the masked image features and noisy latents into separate branches, which increases textual coherence and improves masked image preservation. We show in Fig.~\ref{fig:teaser} that these methods do not perform well for generating reflections on the mirror.

\noindent
\textbf{Reflection in vision tasks.} Reflection has been extensively explored for image enhancement tasks such as single image reflection removal~\cite{wei2019single, hu2023single, levin2004separating}. This task is ill-posed in nature and requires additional priors to be solved. Other methods~\cite{kong2011high,lei2020polarized, li2020reflection} use multiple images to solve this task. Specifically, they use polarization cues to remove reflection from the input image. Further, reflection cues are used to detect the glass/reflective surfaces in the real world~\cite{mei2020don,lin2021rich}. Recently, PromptRR~\cite{wang2024promptrr} uses diffusion models for solving the single image reflection removal task. Further,~\cite{meng2024mirror,zeng2023mirror-nerf,liu2024mirrorgaussian} solves the challenge of reconstructing mirror reflections in a 3D scene for the novel-view synthesis task. In this work, we attempt to solve the challenging task of generating controlled photo-realistic and geometrically consistent mirror reflections for an object in an input image, which has not been addressed in previous works to the best of our knowledge.
\section{~\datasetname{}: A synthetic dataset of mirror reflections}
\label{sec:dataset}

\begin{figure*}[!t]
    \centering
    \includegraphics[width=\linewidth]{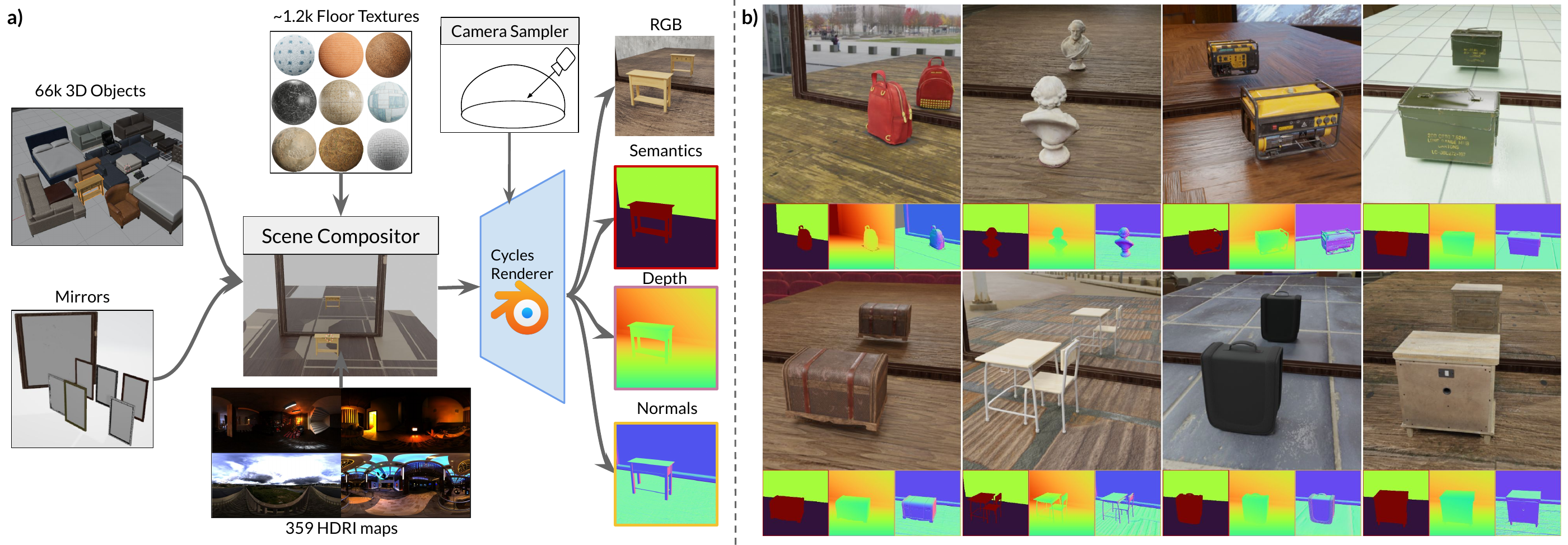}
    \caption{~\datasetname: a) \textbf{Dataset creation pipeline} - We sample diverse 3D objects, mirrors as 2D planes and diverse floor textures to compose a scene in a blender environment. To enhance realism, we sample high-quality HDRI environment maps as backgrounds. We sample cameras from varied viewpoints, capturing the mirror and the object, and use Blender to render RGB images and dense 2D annotations. b) \textbf{Samples from SynMirror} - The generated scenes have complex geometry, textures, and high diversity. The renderings have accurate dense annotations for semantic, depth and normal maps at the original image resolution.}
    \label{fig:dataset_pipeline}
    \vspace{-4mm}
\end{figure*}

We observe that state-of-the-art diffusion models struggle to generate geometrically consistent results for mirror reflections as shown in Fig.~\ref{fig:teaser} and Fig.~\ref{fig:mirror-reflect-sd}. We hypothesize that the cause of this inferior performance is due to the limited number of samples of images with realistic mirror reflections in various existing datasets used to train these models. Further, we find that existing mirror datasets are inadequate for training generative models as they are primarily designed for reflective mirror detection~\cite{Yang_2019_ICCV} and lack object diversity~\cite{mirror3d2021tan}, which is required to incorporate the priors of mirror reflections in diffusion models. To address this, we propose~\datasetname{}, a first-of-its-kind large-scale synthetic dataset on mirror reflections, with diverse \textit{mirror types, objects, camera poses, HDRI backgrounds} and \textit{floor textures}. We provide a comprehensive characteristic comparison of~\datasetname{} with existing mirror datasets in Tab.~\ref{tab:data_characteristic}. Note that~\datasetname{} is more than six times larger than all the existing mirror datasets combined. Further, our data generation pipeline renders color images, instance segmentation masks, depth maps and normal maps as shown in Fig.~\ref{fig:dataset_pipeline}.

Additionally, we create~\testsetname{}, a subset of~\datasetname{}, which serves as a challenging benchmark for generative tasks on mirror reflections.~\datasetname{} can also be leveraged to benchmark other downstream tasks such as monocular depth estimation and novel-view synthesis.

\begin{table}[!t]
 \caption{\textbf{A comparison between~\datasetname{} and other mirror datasets.} The proposed dataset has more attributes and is six times larger in size than all other existing datasets combined.}
\begin{adjustbox}{width=\linewidth}
\begin{tabular}{@{}l|cccc@{}}
\toprule
\multicolumn{1}{c|}{Dataset} & Type &  Size & Attributes\\ \midrule
 MSD~\cite{Yang_2019_ICCV} & Real  & 4018 &  RGB, Masks \\
 Mirror-NeRF~\cite{zeng2023mirror-nerf} & Real \& Synthetic  & 9 scenes &  RGB, Masks, Multi-View\\
 DLSU-OMRS~\cite{DLSU} & Real   & 454 &   RGB, Mask \\
 TROSD~\cite{sun2023trosd} & Real  & 11060  & RGB, Mask\\
 PMD~\cite{PMD:2020} & Real   & 6461 &   RGB, Masks\\
 RGBD-Mirror~\cite{mei2021depth} & Real   & 3049 &   RGB, Depth\\
 Mirror3D~\cite{mirror3d2021tan} & Real   & 7011 &  RGB, Masks, Depth\\ 
 \midrule
\textbf{\datasetname(Ours)} & Synthetic & \textbf{198204} &  RGB, Depth, Masks, Normals, Multi-View\\ \bottomrule
\end{tabular}
\end{adjustbox}
\label{tab:data_characteristic}
\end{table}

\subsection{Dataset Generation and Processing}
\label{subsec:dataset_generation}
\paragraph{Object Source.}~\datasetname{} consists of 3D assets from two widely used 3D object datasets - Objaverse~\cite{objaverse} and Amazon Berkeley Objects (ABO)~\cite{collins2022abo}. Objaverse is a large scale dataset consisting of $800K$ 3D assets with diverse categories and ABO contains catalogued 3D models with complex geometries that correspond to real world household objects. However, some objects from Objaverse are poorly rendered or have low-quality textures. Thus, we use a filtered list of $64K$ 3D objects as filtered by OBJECT 3DIT~\cite{michel2024object}. Despite this initial filtering, some \textit{“spurious”} objects do not show mirror reflections pertaining to specific shader properties of objects. We elaborate on the filtering method to remove \textit{“spurious”} objects in Appendix~\ref{subsec_appendix:filtering_spurious}. Post filtering, we obtain a subset of $58,115$ high-quality 3D assets from Objaverse. 
We include all $7,953$ objects from the ABO dataset to cover a wide range of object shapes and appearances, and thus resulting in $66,068$ total objects.

\noindent
\textbf{Scene Setting.} 
We use a virtual environment in Blender to compose scenes with realistic reflections. We follow a set of heuristic rules to compose a scene with the 3D asset, a floor, and a mirror; an example is shown in Fig.~\ref{fig:dataset_pipeline}. We model mirrors as rectangular planes of varying sizes and frame textures. The floor is modeled as a plane with a diverse set of textures sampled from $1182$ CC-textures~\cite{Denninger2023}. 

We compose a scene, by first placing a mirror vertically at a fixed location. Next, we define a region at a fixed distance to the mirror where the 3D object will be placed, represented as a unit cube. We normalize the object to fit in the unit cube. We also rotate the 3D object around its $y-axis$ to increase diversity in object poses. For modeling the background, we use $359$ high-quality HDRI environment maps from PolyHeaven~\cite{polyhaven}. We categorize the floor textures and HDRI maps into \textit{indoor} and \textit{outdoor} categories to simulate realistic \textit{indoor}/\textit{outdoor} scenes enhancing photorealism of the renderings. For illuminating the scene, we use an area light placed slightly above and behind the object at a $45$ degree angle pointing towards the object and the mirror.

\noindent
\textbf{Rendering.}
We te a pool of $19$ camera poses by interpolating between two extreme camera poses while ensuring that the object and its reflection in the mirror are visible. For each scene iteration, we randomly sample 3 camera poses and render the scene from these virtual cameras by leveraging BlenderProc~\cite{Denninger2023}, to obtain RGB, depth maps, surface normal maps, and semantic labels. To obtain high-quality photo-realistic renderings, we render at a $512 \times 512$ resolution using 1024 cycles from Blender's Cycles renderer. This allows us to create a rich and comprehensively annotated~\datasetname{} dataset for studying a variety of mirror related tasks. More details about the dataset are provided in Appendix~\ref{appendix:dataset_related}.

\section{Method}
\label{sec:method}

\begin{figure*}
    \centering
    \includegraphics[width=\linewidth]{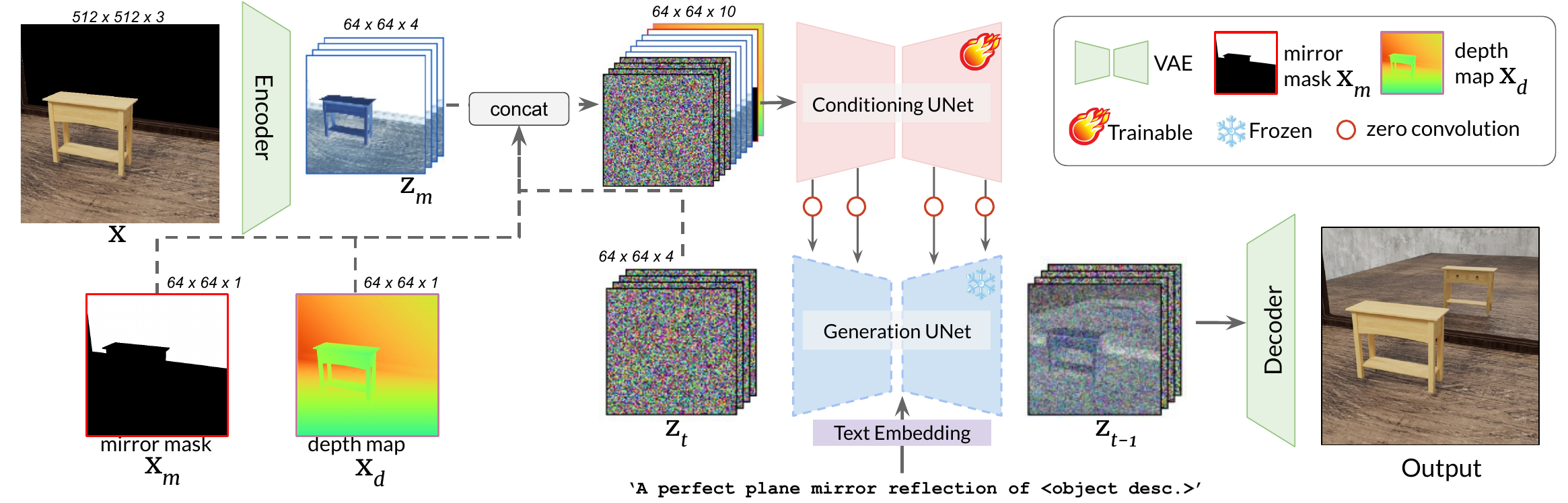}
    \caption{\textbf{Overview of the architecture.} We encode the input image \textbf{$x$} using a pre-trained image encoder from Stable Diffusion to get \textbf{$z_m$}. Subsequently, we resize the mirror mask \textbf{$m$} and depth map \textbf{$d$} to obtain resized mask \textbf{$x_m$} and depth \textbf{$x_d$}. Then, we concatenate noisy latents \textbf{$z_t$}, \textbf{$z_m$}, \textbf{$x_m$} and \textbf{$x_d$} which are fed into the Conditioning U-Net \textbf{$\epsilon^{'}_{\theta}$}. Each layer of the Generation U-Net \textbf{$\epsilon_{\theta}$} is conditioned via zero convolutions with corresponding layers of \textbf{$\epsilon^{'}_{\theta}$}. Additionally, \textbf{$\epsilon_{\theta}$} is conditioned by text embeddings. The pre-trained decoder then decodes the denoised latent to produce an image with mirror reflections. Detailed information can be found in Sec.~\ref{subsec:method}}
    \label{fig:architecture}
\end{figure*}

We briefly introduce diffusion models in Sec.~\ref{subsec:prelims}. Then, we present our method~\methodname{} in Sec.~\ref{subsec:method}. Fig.~\ref{fig:architecture} provides an overview of our method. 

\subsection{Preliminaries}
\label{subsec:prelims}
Diffusion models are a family of generative models that generate images by iterative denoising. In the forward diffusion process, a Gaussian noise $\epsilon \sim \mathcal{N} \left( 0,1 \right)$; is sequentially added for $T$ timesteps to a clean sample $x_0$ to get a noisy sample $x_T$. In the backward diffusion process, a clean image $x_0$ is generated by iterative denoising of noisy image $x_T$. The iterative denoising process is modeled with a denoising network $\epsilon_{\theta}$ conditioned on the timestep $t\in\{1,T\}$ and optional conditioning $c$ (e.g. text prompts, inpainting masks). The denoiser is trained with simple mean square loss $L_{DM}$ as follows:
\vspace{-2mm}
\begin{equation}
L_{DM} = \;\; E_{x_0,\epsilon \sim \mathcal{N} \left( 0,I \right) ,t} ||\epsilon -\epsilon _{\theta}\left(z_t, t, c\right)||^2 
 \label{eq:train_objective}
\end{equation}

\noindent 
Training diffusion models directly on the large resolution images $x_0$ is computationally demanding as it needs several denoising steps to generate a single image. Latent Diffusion Models~\cite{rombach2022high} propose to apply a diffusion process in a compressed latent space of a pre-trained Variational Autoencoder. This enables efficient training and fast inference for generating large-resolution images.





\subsection{MirrorFusion}
\label{subsec:method}
Though trained on large-scale datasets, existing state-of-the-art diffusion models fail to generate consistent reflections with accurate object shape and scene appearance as shown in Fig.~\ref{fig:teaser}. We propose~\methodname{}, a novel framework for generating accurate mirror reflections by formulating it as an inpainting problem. Given an input scene image and a mirror mask,~\methodname{} fills the masked region with the consistent reflection of the object and the scene. Generating accurate reflections requires a precise 3D understanding of the scene to reason about the distance of objects from the mirror and the shapes of objects. Hence, modeling reflections with just a 2D inpainting model is suboptimal, and we need to inject explicit 3D cues during the inpainting process. Thus, we propose to condition the inpainting approach with depth maps. The geometric signal from the depth map enables us to generate accurate and consistent reflections that adhere to the 3D structure of the scene and the object. Fig.~\ref{fig:architecture} shows the overview of our method. 


\subsubsection{Model architecture}
\label{subsec:depth_mask_guidance}
\methodname{} is a diffusion-based inpainting model conditioned on the input mirror mask and depth map. We use a base dual branch architecture for inpainting following BrushNet~\cite{ju2024brushnet} as shown in Fig.~\ref{fig:architecture}. The core idea is to clone a pretrained diffusion model $\epsilon_{\theta}$ without cross-attention layers to $\epsilon^{'}_{\theta}$. Subsequently, the features from the conditioning model $\epsilon^{'}_{\theta}$ are inserted into the generation model $\epsilon_{\theta}$ using zero-convolutional layers. During training, \textit{only the conditioning model is updated, keeping the generation model frozen}. This conditioning mechanism provides a strong hierarchical conditioning for generation without altering the original generation model.

\vspace{-2mm}
\subsubsection{Depth conditioning}
\vspace{-2mm}
Geometric information about objects and scenes is crucial for generating 3D consistent reflections. Recent works~\cite{pandey2024diffusion,bhat2024loosecontrol} show that injecting depth maps enables 3D geometric control in the diffusion models. 
Inspired by this, we utilize depth-conditioning for our inpainting architecture. Specifically, the noisy latent $z_t$, masked image latent $z_m$, inpainting mask $x_m$, and the depth map $x_d$ are all concatenated and passed as input to the conditioning U-Net ${\epsilon^{'}}$. The generation U-Net ${\epsilon}$ is an unaltered text-to-image diffusion model, which takes a noisy latent $z_t$ and predicts a cleaner version $z_{t-1}$. Each layer of generation U-Net ${\epsilon_i}$ is conditioned with the corresponding layer of conditioning U-Net using zero-convolutions ($\mathcal{Z}$) as follows:

\vspace{-4mm} 
\begin{equation}
 \epsilon _{\theta}\left( z_t,t,c \right) _i=\epsilon _{\theta}\left( z_t,t,c \right) _i+  
 w\cdot\mathcal{Z} \left( \epsilon _{\theta}^{'}\left( \left[ z_t,z_m,x_m,x_d \right] ,t \right) _i \right) 
 \label{eq:conditioning_eq}
\end{equation}

\noindent $w$ is the preservation scale to adjust the influence of conditioning. We set $w$ to be $1.0$ for all our experiments. \\

\begin{figure}[!t]
    \centering
    \includegraphics[width=\linewidth]{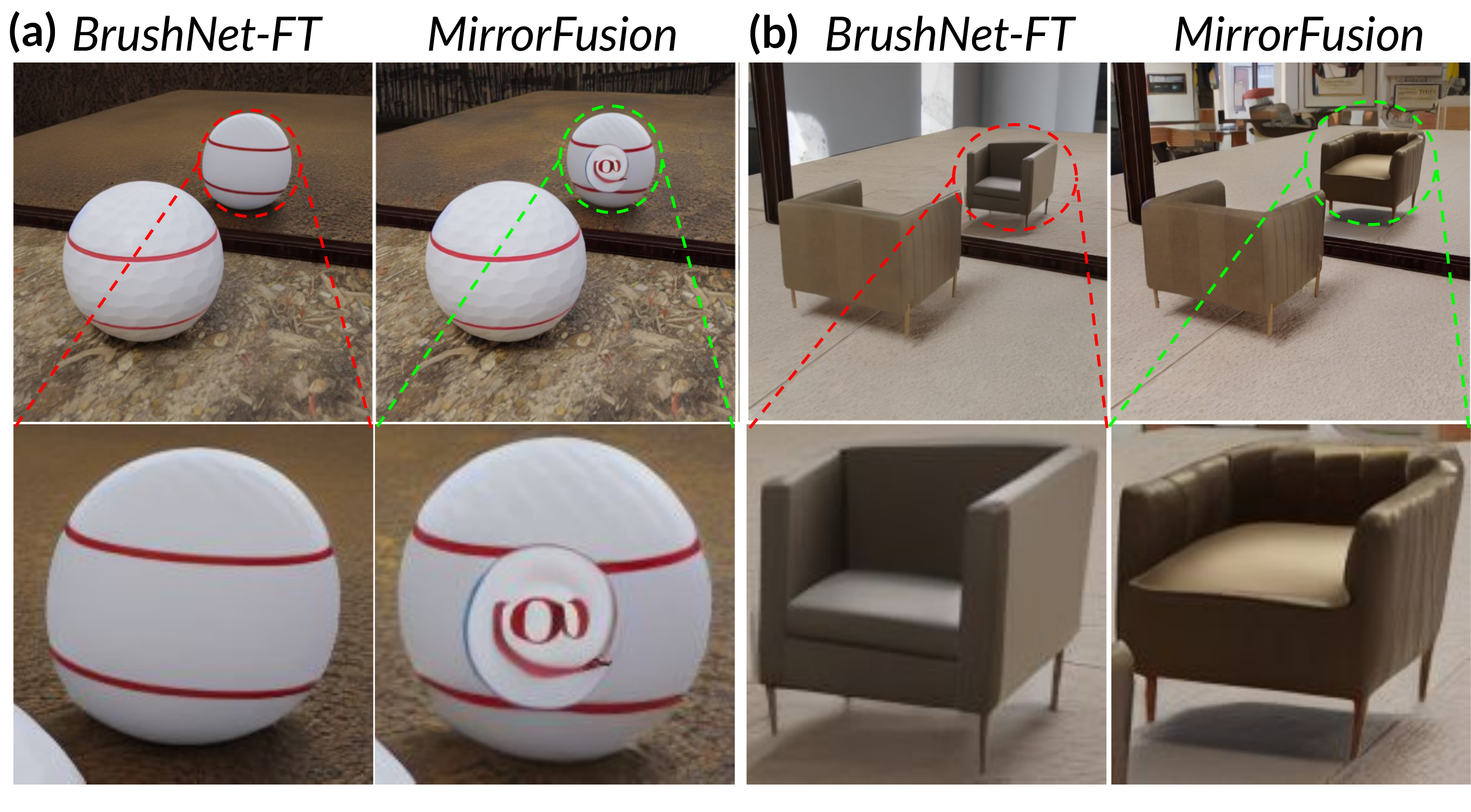}
    \caption{\textbf{Impact of depth conditioning} on the reflection generation quality. Notice the irregular shape of the ``baseball'' and ``chair'' marked in \textcolor{red}{red}. In comparison, our method preserves the structure of the object (marked in \textcolor{green}{green}).}
    \label{fig:depth_significance}
    \vspace{-6mm}
\end{figure}

\noindent\textbf{Impact of Depth Conditioning.}
\label{subsec:depth_significance}
We demonstrate the importance of depth conditioning for the reflection generation task as shown in Fig.~\ref{fig:depth_significance}. 
From Fig.~\ref{fig:depth_significance}~(a), it can be clearly seen that BrushNet~\cite{ju2024brushnet} fine-tuned on~\datasetname{} fails to generate accurate mirror reflection of the object in the input image with high fidelity. For a simple object like a ``baseball ball'', the \textit{``w/o depth''} BrushNet-FT model generates a ball in which the shape is not preserved. 
Similarly, in Fig.~\ref{fig:depth_significance}~(b), the shape of ``chair'' is asymmetrical. These examples show that depth information provided with the proposed normalization scheme generates better reflections on the mirror. 

\vspace{-2mm} 
\noindent
\subsubsection{Depth Normalization} 
\vspace{-2mm}
The range of input depth is between $[0,\infty)$. The encoder of the U-Net expects the input to be in the range $[-1,1]$. Hence, we need to normalize the input depth. As discussed in~\cite{ke2024repurposing}, using affine-invariant depth scaling can bring the input depth into the desired range. Image reflection tasks only require the relative distance between the mirror and the scene it reflects. Depth values behind the mirror will not be critical for the reflection generation task. Hence, we use a specifically tailored normalization for our task which is computed as:
\begin{equation}
 \hat{d} = \left(\frac{d_{clipped}}{d_{max} + \Delta_{depth}} - 0.5\right) \times 2,
\end{equation}
where $d_{clipped}$ is input depth clipped between range $[0,d_{max} + \Delta_{depth}]$. $d_{max}$ represents the maximum depth on the mirror mask $m$. We set $\Delta_{depth}$ to the value of $0.5$. This normalization aids us in adapting to any of the pre-trained monocular depth estimation methods~\cite{ke2024repurposing,yang2024depth}.

\noindent
\textbf{Inference.}
During inference, we assume a predefined mask is available to indicate where the mirror reflection should be generated. The user can easily create this mask, giving better control over the generation of the reflections. We leverage Marigold~\cite{ke2024repurposing}, a monocular depth estimation method to generate the scene depth map. Then, we feed these inputs to our pipeline as shown in Fig.~\ref{fig:architecture}. We show in Appendix~\ref{subsec_appendix:depth_algo_choice} that we can utilize alternative methods, such as DepthAnything~\cite{yang2024depth} as a preferred monocular depth estimation technique, demonstrating the robustness of our method to the choice of different monocular depth estimation techniques. 

\section{Experiments \& Results}
\label{sec:exp_results}
\noindent
In this section, we discuss the dataset, baseline comparisons, and extensive experiments used to evaluate our model. We provide additional training and implementation details in Appendix~\ref{appendix:implementation_details}.

\vspace{1mm}
\noindent 
\textbf{Dataset.}
As discussed in Sec.~\ref{subsec:dataset_generation},~\datasetname{} consists of $66,068$ objects and $198,204$ rendered images. We sample $1000$ objects from the full dataset to create ~\testsetname{}, a benchmark to evaluate our method and various other baselines. In this benchmark, we have $1497$ \textit{``known''} class samples, i.e., these object categories were seen during training, and $1494$ \textit{``unknown''} class samples, i.e., these object categories were unseen during training.~\testsetname{} thus comprises of a total of $2991$ images. We also show the generalization capabilities of our method on a few samples from the Google Scanned Objects(GSO)~\cite{downs2022google} dataset.

\begin{figure}[!t]
    \centering
    \includegraphics[width=\linewidth]{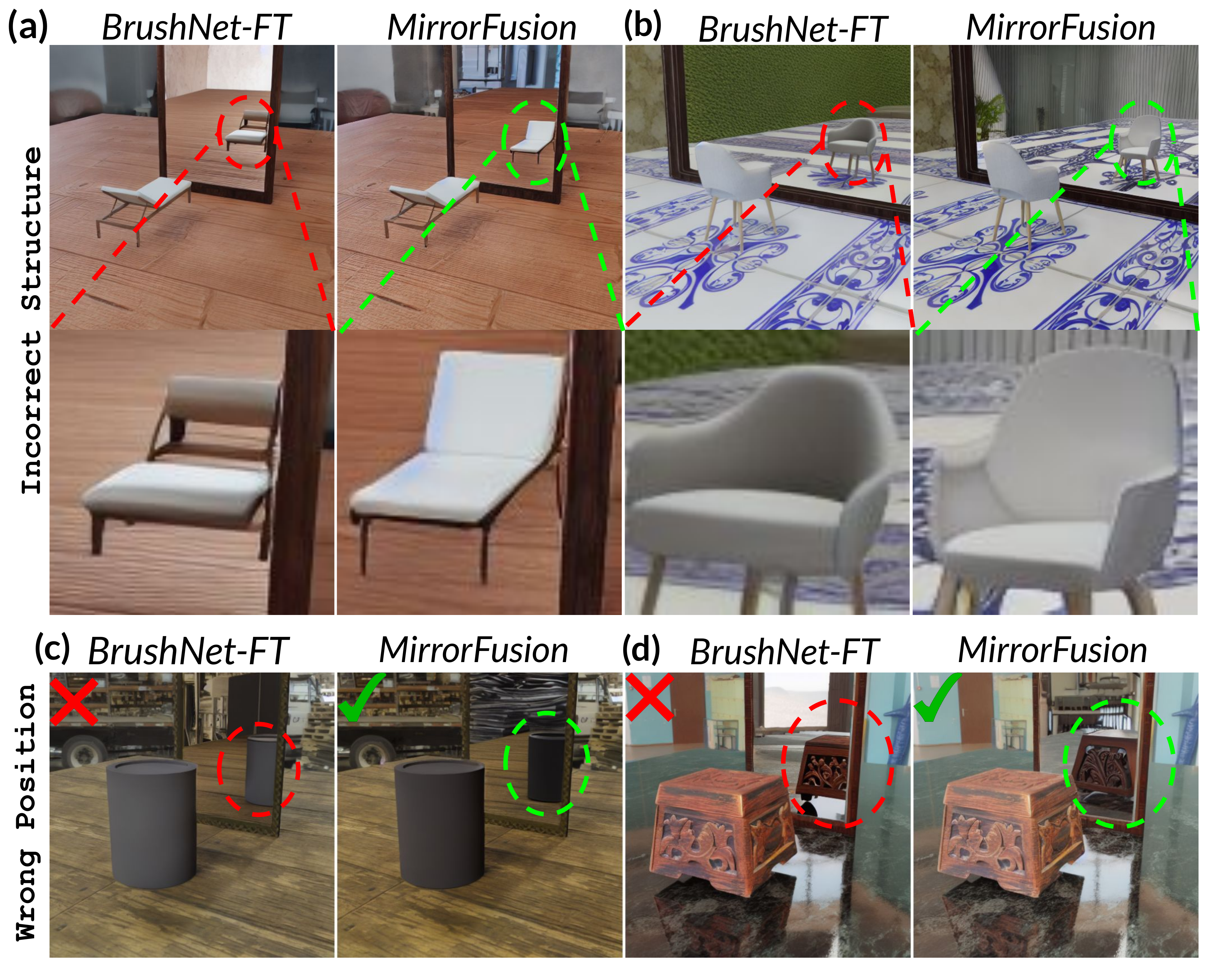}
    \caption{\textbf{Additional Results.} Our method effectively preserves the shape of objects, as demonstrated in (a) the lawn chair and (b) the swivel chair. Check in the zoomed-in regions. Additionally, our method accurately positions the objects within the mirror (c) and (d), corroborating the effectiveness of depth-conditioning in our method. Text-prompts used are described in Appendix~\ref{subsec_appendix:captions_used_in_exp}.}
    \label{fig:our_qual_results}
    \vspace{-4mm}
\end{figure}

\vspace{1mm}
\noindent
\textbf{Baselines.}
As discussed in Sec.~\ref{sec:method}, we formulate generating reflection of an object as an image-inpainting problem. We evaluate various state-of-the-art inpainting methods on~\testsetname{}. We compare our method with pre-trained Stable-Diffusion-Inpainting~\cite{rombach2022high}, PowerPaint~\cite{zhuang2023task} and BrushNet~\cite{ju2024brushnet}. We denote zero-shot methods by appending ``\textbf{-ZS}'' to their names. We fine-tune BrushNet on ~\datasetname{} and refer to this fine-tuned version as \textbf{``BrushNet-FT''} hereafter.

\begin{figure*}[!t]
    \centering
    \includegraphics[width=\linewidth]{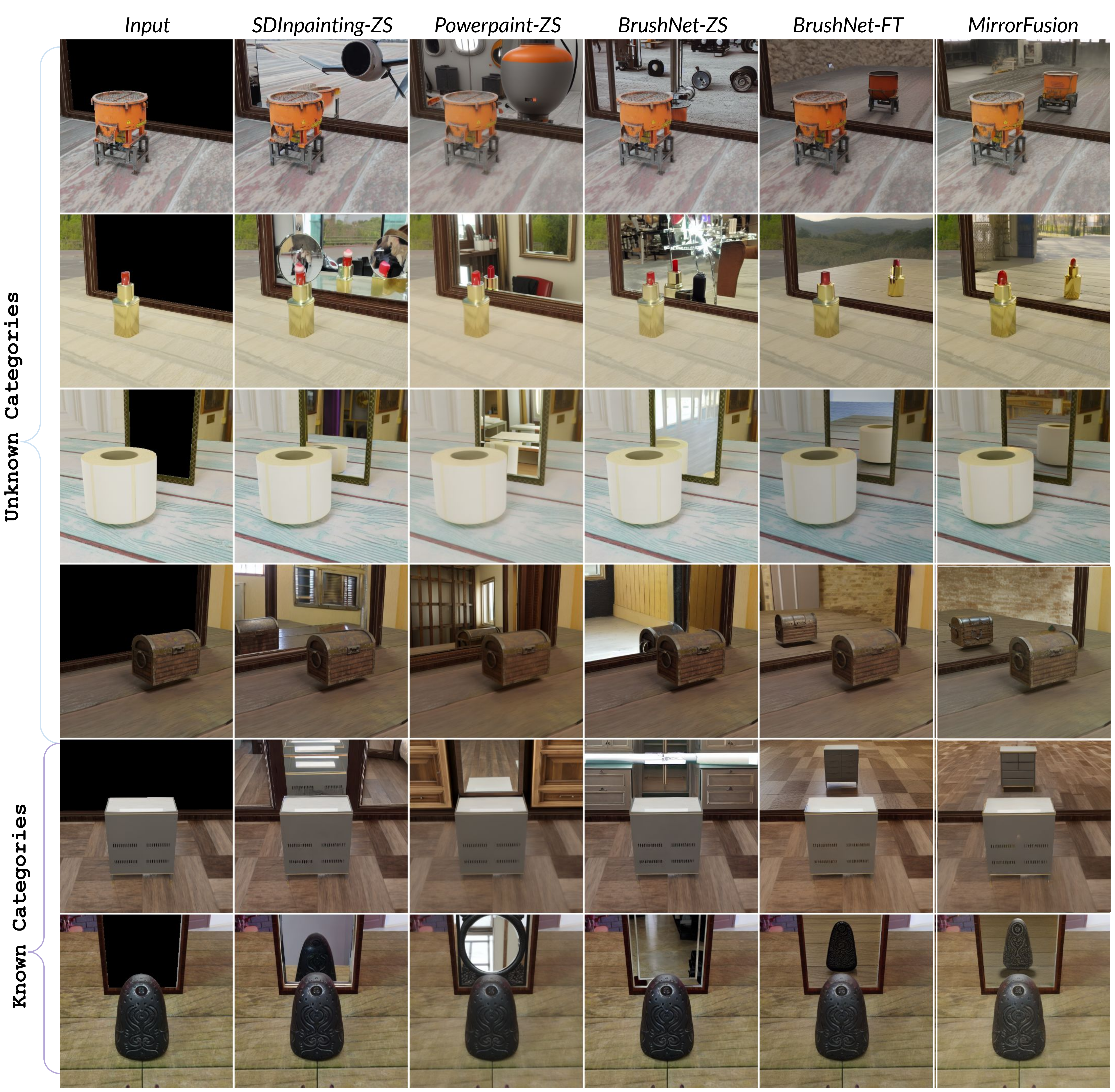}
    \caption{\textbf{Reflection generation comparison with general inpainting methods.} We compare our results with zero-shot baselines Stable Diffusion 1.5 Inpainting-ZS, PowerPaint-ZS and BrushNet-ZS. Further, we finetune BrushNet on~\datasetname{} and refer to it as BrushNet-FT. The top four rows compare results on \textit{``unknown''} categories, and the bottom two rows show results on \textit{``known''} categories from~\testsetname{}. Zero-shot methods either fail to generate a reflection on the mirror or generate a reflection at an incorrect position. In comparison, BrushNet-FT generates plausible reflections, but with geometric inaccuracies. Our method improves on shape preservation of the object, floor texture and correct placement of the object in the mirror reflection.}
    \label{fig:cmp_qual}
\end{figure*}

\noindent
\textbf{Metrics.}
We benchmark based on four aspects: masked region preservation, reflection generation quality, Reflection Geometry and text alignment.
\begin{itemize}
 
    \item \textbf{\textit{Masked Region Preservation.}} We report Peak-Signal-to-Noise ratio (PSNR), Structural Similarity (SSIM) and Learned Perceptual Image Patch Similarity (LPIPS)~\cite{zhang2018unreasonable} in the unmasked region between generated and the real image. This shows how much original image content is preserved by an inpainting method.
    \item \textbf{\textit{Reflection Generation Quality.}} 

    For measuring the quality of the reflection of the object and the scene, we compute PSNR, SSIM and LPIPS on the masked region containing the object and floor of the ground truth image.
      
    \item \textbf{\textit{Reflection Geometry.}} We measure the geometric accuracy of the generated reflection using Intersection over Union (IoU) between the segmentation mask of the ground-truth object and the generated object in the reflection region specified by the input mirror mask. We utilize SAM~\cite{kirillov2023segment} to get the mask of an object in the reflection region. More details are provided in Appendix~\ref{subsec_appendix:seg_masks}.
    \item \textbf{\textit{Text Alignment.}} To evaluate the text-image consistency between the generated image and the text prompts, we use CLIP~\cite{radford2021learning} Similarity. 
\end{itemize}

\subsection{Qualitative Results}
\label{subsec:qual_results}
\paragraph{Comparison with Zero-shot Baselines.} 
We observe that all zero-shot baselines fail in generating realistic reflections on the mirror. PowerPaint generates the ``lipstick'' at the incorrect position, whereas Stable Diffusion 1.5 Inpainting generates two reflections when only one is present. No zero-shot method is able to provide a plausible reflection for a ``cement mixer'' as shown in Fig.~\ref{fig:cmp_qual} (top row). BrushNet-FT performs better than the zero-shot methods, which shows the utility of the proposed dataset. However, it has issues such as the incorrect size of the object in the generated reflections for ``cement mixer'' and ``lipstick'' as shown in the first two rows of Fig.~\ref{fig:cmp_qual}. In comparison, our method is able to generate realistic reflections of the objects.

\vspace{2mm}
\noindent
\textbf{Additional Results.}
Fig.~\ref{fig:our_qual_results} shows more comparisons between BrushNet-FT and our method. First row shows that there are some structural inaccuracies in the generated reflection. Check how BrushNet-FT is not able to get the structure of the ``lawn chair'' in Fig.~\ref{fig:our_qual_results}~(a) and swivel chair in Fig.~\ref{fig:our_qual_results}~(b). However, our method is able to generate the reflection of the object in a geometrically accurate position. Further in Fig.~\ref{fig:our_qual_results}~(c) and (d), notice that the reflection is generated at wrong position in the mirror by BrushNet-FT.  

\begin{figure}
    \centering
    \includegraphics[width=\linewidth]{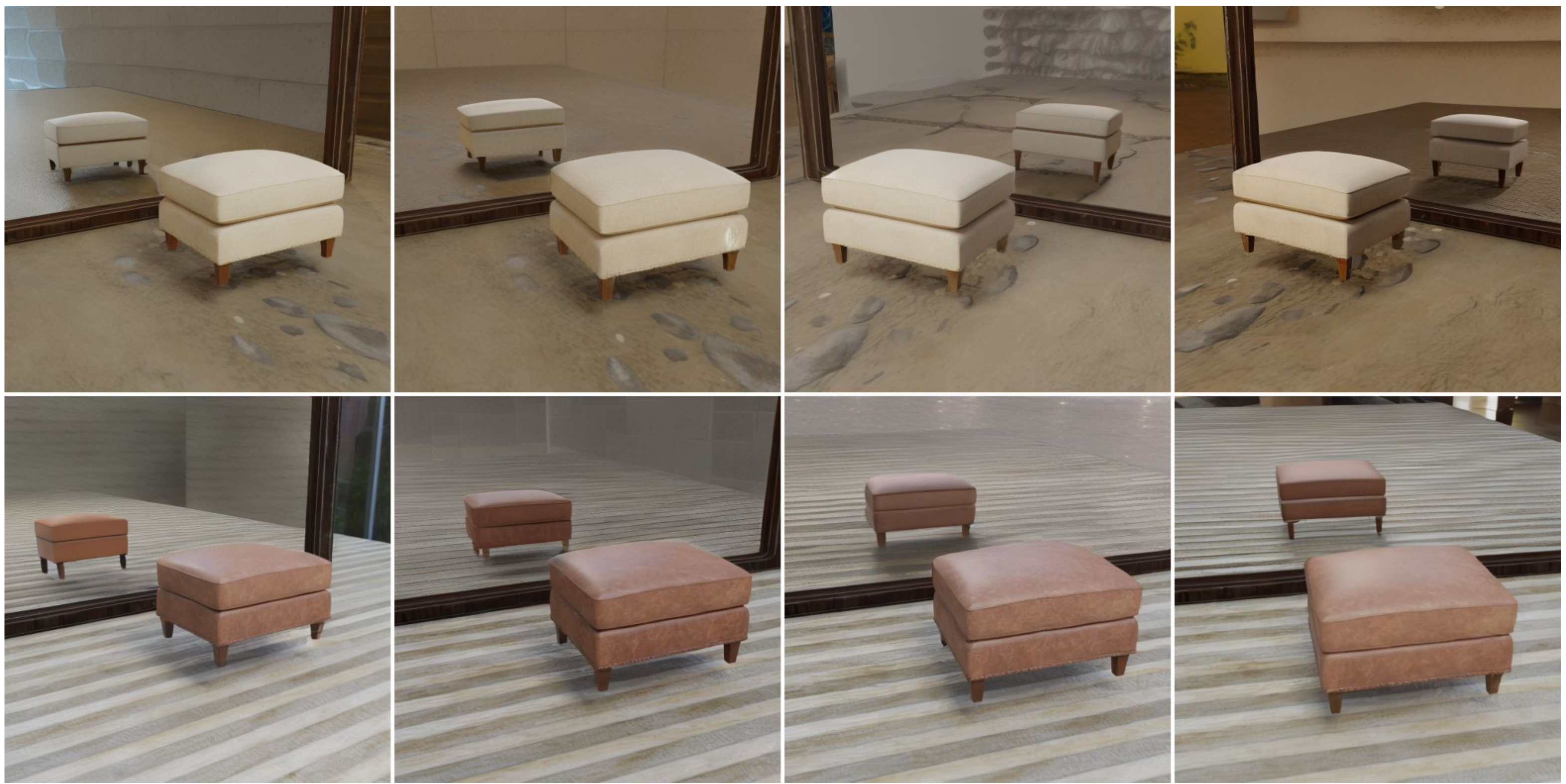}
    \caption{\textbf{Change of Viewpoints for mirror and object.}
    Our method preserves the shape of the object from different viewpoints. This illustrates our method's ability to utilize 3D cues and generate accurate reflection of the object.}
    \label{fig:novel-views}
\end{figure}

\begin{figure}[!t]
    \centering
    \includegraphics[width=\linewidth]{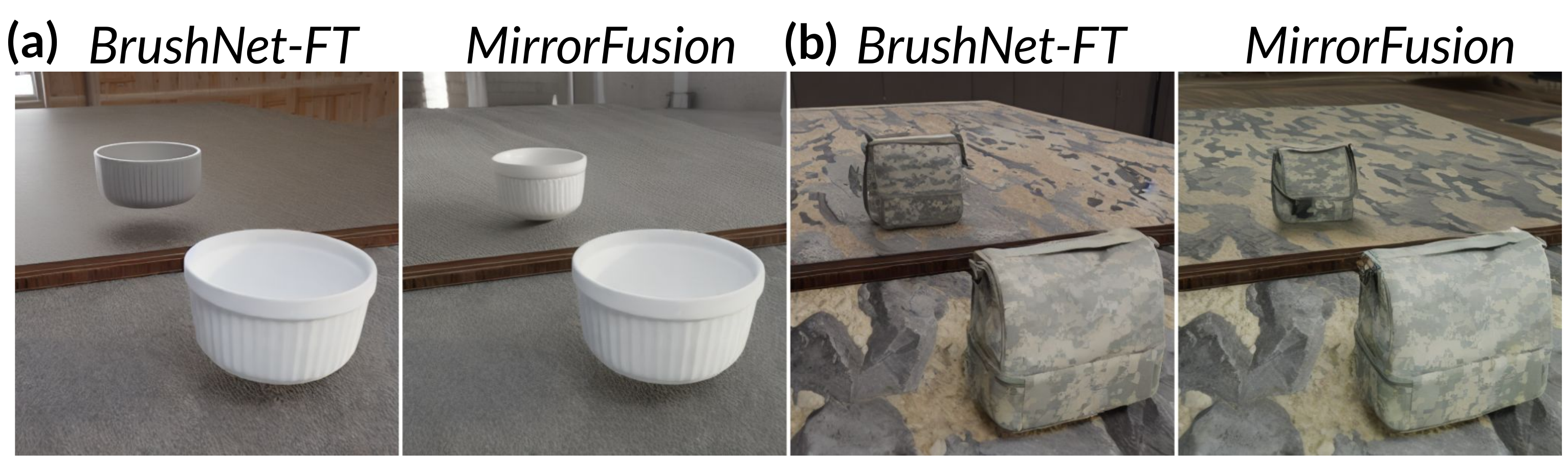}
    \caption{\textbf{Generalization on GSO~\cite{downs2022google}.} Our method generates accurate reflections for unseen real-world scanned objects. This substantiates the generalization capabilities of our method.}
    \vspace{-5mm}
    \label{fig:wild_gso} 
\end{figure}

\vspace{2mm}
\noindent
\textbf{Qualitative Results on GSO.}
We further benchmark the performance of our method on a completely held-out set of GSO objects. Fig.~\ref{fig:wild_gso} compares our method with the fine-tuned baseline: BrushNet-FT. Notice that the bowl is floating in the air for BrushNet-FT and the size of the bag in the reflection is large and unnatural. In comparison, our method generates the reflection with better photo-realism and geometric accuracy. 

\vspace{2mm}
\noindent
\textbf{Change of Viewpoints.} 
To evaluate the consistency of our method in generating reflections across varying viewpoints, we designed a continuous trajectory for testing. The results inferred from our method, as depicted in Fig.~\ref{fig:novel-views}, demonstrate that the reflection of the ``sofa-seat'' remains consistent as the viewpoint shifts. Additionally, our method preserves high-fidelity reflections for the floor.

\subsection{Ablation studies}
\label{subsec:quant_results}

\begin{table}[!t]
\centering
\caption{\textbf{Image Generation Quality.} We compare the quality of the inpainted image with fine-tuned baseline method. The best results are shown in \textbf{bold}. Our method outperforms the baseline across all metrics, proving its effectiveness.}
\label{tab:quant_iq}
\begin{adjustbox}{width=\linewidth}
\begin{tabular}{@{}c|ccc|c@{}}
\toprule
Metrics & \multicolumn{3}{c|}{Masked Image Preservation}  & Text Alignment \\ \midrule
Models & \textbf{PSNR} $\uparrow$ & \textbf{SSIM} $\uparrow$ & \textbf{LPIPS} $\downarrow$  & \textbf{CLIP Sim} $\uparrow$ \\ \midrule
\textbf{Brushnet-FT}~\cite{ju2024brushnet} & 23.06 & \textbf{0.84} & 0.058 & 24.90 \\
 \textbf{Ours} & \textbf{24.22} & \textbf{0.84} & \textbf{0.051} & \textbf{25.23} 
 \\ \bottomrule
\end{tabular}
\end{adjustbox}
\end{table}


\begin{table}[!t]
\centering
\caption{\textbf{Reflection Quality.} We compare the quality of the generated reflection image with the baseline method. We observe that our method has better object quality metrics. Best results are shown in \textbf{bold}.}
\vspace{-2mm}
\begin{adjustbox}{width=\linewidth}
\begin{tabular}{@{}c|ccc|c@{}}
\toprule
Metrics & \multicolumn{3}{c|}{Reflection Generation Quality}  & Reflection Geometry \\ \midrule
Models & \textbf{PSNR} $\uparrow$ & \textbf{SSIM} $\uparrow$ & \textbf{LPIPS} $\downarrow$  & \textbf{IoU} $\uparrow$ \\ \midrule
\textbf{Brushnet-FT}~\cite{ju2024brushnet} & 19.15 & \textbf{0.84} & 0.082 & 0.566 \\
 \textbf{Ours} & \textbf{20.35} & \textbf{0.84} & \textbf{0.075} & \textbf{0.567} \\ \bottomrule
\end{tabular}
\end{adjustbox}
\label{tab:quant_reflection}
\end{table}


Tab.~\ref{tab:quant_iq}~\&~\ref{tab:quant_reflection} quantitatively compare BrushNet-FT and~\methodname{} on the image quality and the generated reflection quality. These values are reported on~\testsetname{}. We generate $4$ outputs for each test sample using different random seeds. We then select the image with the best mask SSIM score as the representative image out of the four images. The reported value for any metric is the average of that metric for all representative images across the dataset.~\methodname{} (Ours) with depth cues outperforms BrushNet-FT, which doesn't take depth as an input. This corroborates the necessity of adding depth as an input to the model. We discuss limitations and societal impact in Appendix~\ref{appendix:limitations}. 

\section{Conclusion}
In this work, we propose~\datasetname{}, a large-scale scale challenging and diverse dataset to train generative models for the task of generating realistic mirror reflections. We identify shortcomings in current models and propose~\methodname{}, a novel inpainting method conditioned on depth maps for generating geometrically accurate mirror reflections. 
Extensive qualitative and quantitative results on~\testsetname{} shows the superior performance of~\methodname{} in comparison to other methods. Our work is the first step towards generating geometrically accurate and plausible mirror reflections using diffusion based generative models. We believe that~\datasetname{} and~\testsetname{} will pave way for research in several mirror-related tasks. 

\noindent \textbf{Acknowledgement.} We are grateful to Kotak IISc AI-ML Centre for providing the compute resources. We thank Om Rastogi for setting up the metric evaluation framework. 

\newpage

{
    \small
    \bibliographystyle{ieeenat_fullname}
    \bibliography{main}
}

\newpage
\clearpage
\maketitlesupplementary

\appendix

\tableofcontents

\addtocontents{toc}{\protect\setcounter{tocdepth}{2}}


\section{Dataset}
\label{appendix:dataset_related}

\begin{figure*}[!t]
    \centering
    \includegraphics[width=0.925\linewidth]{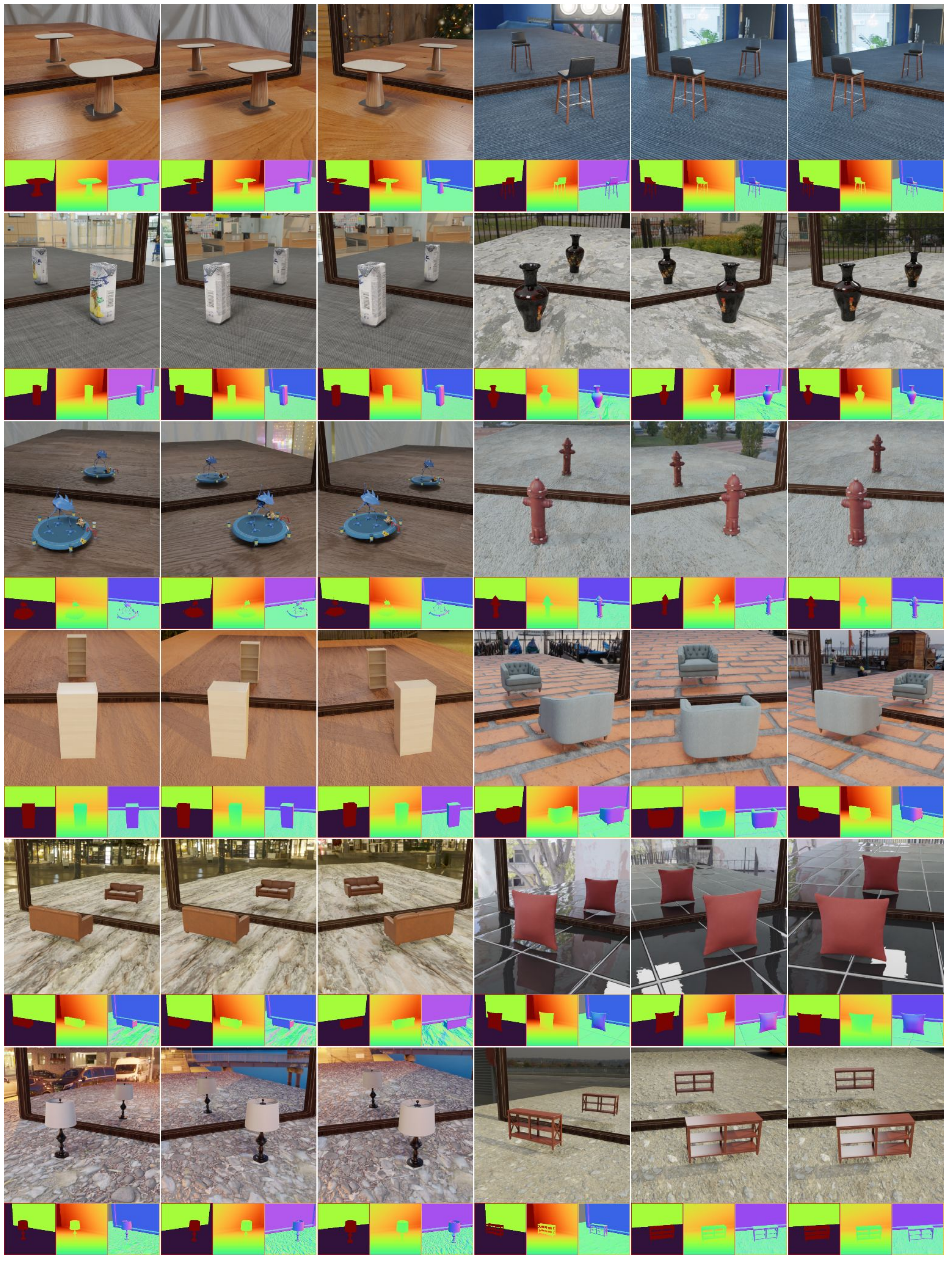}
    \caption{\textbf{Samples from~\datasetname{}.} }
    \label{fig:supp_full_wall_mirrors}
\end{figure*}

\begin{figure*}[!t]
    \centering
    \includegraphics[width=0.925\linewidth]{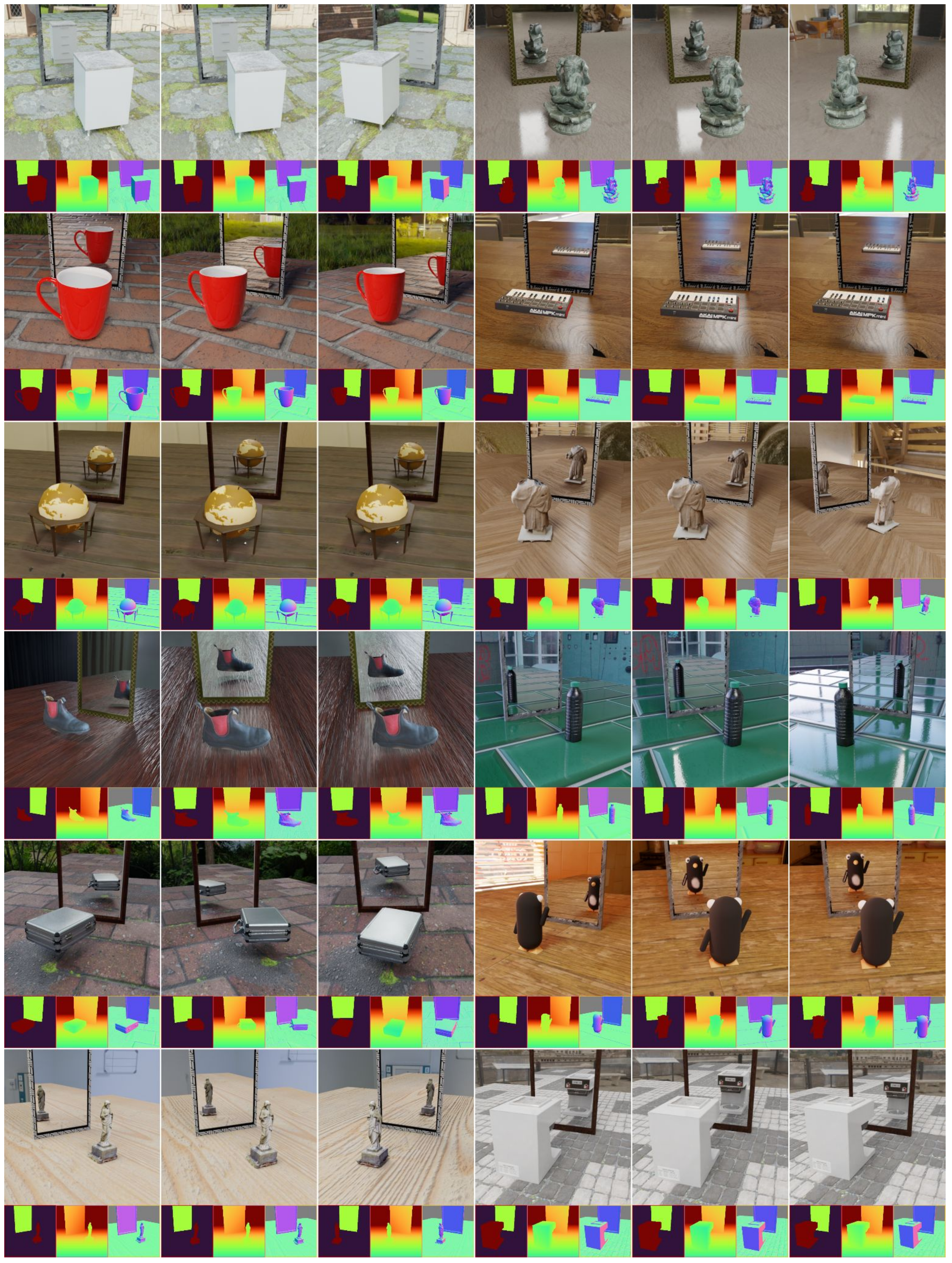}
    \caption{\textbf{\textbf{Samples from~\datasetname{}}.} }
    \label{fig:supp_small_mirrors}
\end{figure*}

Our dataset consists of $198,204$ rendered images from $66,068$ objects: $58,115$ objects from Objaverse~\cite{objaverse} and $7,953$ from the ABO~\cite{collins2022abo} dataset. We utilize captions provided by Cap3D~\cite{cap3d} during training. We provide more details in Sec.~\ref{sec:dataset}. To illustrate the diversity in 3D objects, floor textures and HDRI backgrounds, we present more samples in Fig.~\ref{fig:supp_full_wall_mirrors} and~\ref{fig:supp_small_mirrors}. 


\subsection{Filtering out Spurious objects}
\label{subsec_appendix:filtering_spurious}
We discuss how we filter 3D objects from Objaverse~\cite{objaverse} and Amazon Berkeley Objects (ABO)~\cite{collins2022abo} in Sec.~\ref{subsec:dataset_generation}. In spite of the initial filtering, we observe some ``spurious'' objects, for which the reflection is not visible in the mirror. Algorithm~\ref{algo:spurious} illustrates the pseudo-code to identify such ``spurious'' objects. Specifically, using Blender's Python API, we check the material property of each \textit{child} in the input mesh $\mathcal{M}$ of a 3D object. We expect the 3D objects to be in standard 3D formats: \textit{``*.glb, *.gltf, *.obj, *.fbx''}. If any node in the material property has the attributes: ``Mix-Shader'', and the name of the input to this node is ``Fac,'' and the name of the linked node is ``Light Path'', then we observe that the reflection of such a 3D model does not appear in the mirror. We prune out such objects from the initial filtered list. The new filtered list will be made public along with the dataset for future research.

\begin{algorithm}[!t]
 \caption{Determine if a 3D Object is Spurious}
 \label{algo:spurious}
		 \textbf{Input}: A 3D model $\mathcal{M}$ \\
		 \textbf{Output}: True, if a 3D model is spurious,  else False
		 \begin{algorithmic}[1]
			\For{$\mathcal{C} \gets$ child $\in \mathcal{M}$}
                    \For{$\mathcal{T} \gets$ material $\in \mathcal{C}$}
                            \For{$\mathcal{N} \gets$ node $\in \mathcal{T}$.material}
        \If{ ($\mathcal{N}$.name $==$ ``Mix-Shader'') \\ \indent \indent \indent \,\,\textbf{and}  ($\mathcal{N}$.input.name $==$ ``Fac'') \\ \indent \indent \indent \,\,\textbf{and} ($\mathcal{N}$.linked.name $==$ ``Light Path'')}
        \State \Call{Return}{True}
    \EndIf
            \EndFor
        \EndFor
    \EndFor
     \end{algorithmic}
\end{algorithm}
\vspace{-2mm}
\subsection{Preparation of~\testsetname{}}
\label{subsec_appendix:benchmark_details}
\begin{figure}[!t]
    \centering
    \includegraphics[width=\linewidth]{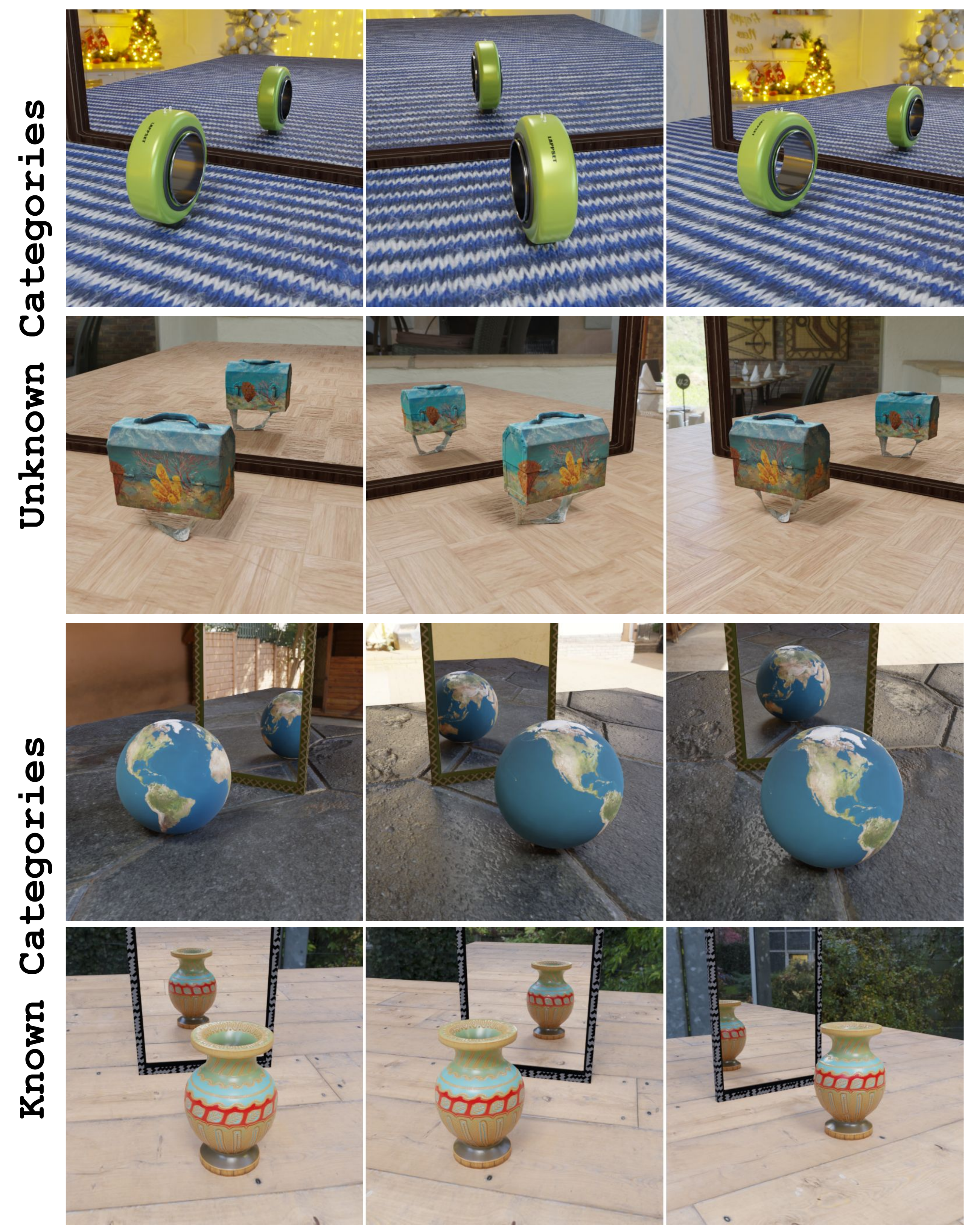}
    \caption{\textbf{Samples from~\testsetname{}.}  The first two rows contain samples from \textbf{``Unknown''} categories and the bottom two rows contain samples from \textbf{``Known''} categories. Notice the challenging nature of~\testsetname{}. We provide more details in Appendix~\ref{subsec_appendix:benchmark_details}}
    \label{fig:supp_mirr_benchmark}
\end{figure}

~\testsetname{} aims to benchmark various generative models at the task of generating perfect mirror reflections.~\testsetname{} is created by sampling around $1,000$ objects from~\datasetname{}, with 3 rendered samples per object, totalling to $2,991$ samples. Fig.~\ref{fig:supp_mirr_benchmark} shows samples of~\testsetname{}, which consist of two types: 
\begin{enumerate}
    \item \textbf{``Unknown''} class objects, referring to categories not present in the training set. We take the first $500$ objects from Objaverse in \textbf{``Unknown''} category, sorted in the increasing order of category frequency and keep the remaining categories in the training set as \textbf{``Known''} categories. There are $1494$ samples generated from the objects of \textbf{``Unknown''} category.
    \item \textbf{``Known''} class objects, referring to categories included in the training set. There are $1497$ images from this category. This includes renderings from around $250$ objects from Objaverse and around $250$ objects from ABO.
\end{enumerate}

\FloatBarrier

\section{Implementation Details}
\label{appendix:implementation_details}

\subsection{Training Details:~\methodname{}}
\label{subsec_appendix:our_training}
We follow the BrushNet~\cite{ju2024brushnet} architecture for~\methodname{} and provide depth conditioning as discussed in Sec.~\ref{subsec:method}. The Generation and Conditional U-Net weights are initialized from the Stable Diffusion v1.5~\cite{rombach2022high} checkpoint. During training, the weights of the generation U-Net are kept frozen, while the weights of the conditioning network are updated. The extra channels processing the down-sampled depth and mask images in the first convolution layer of the conditioning U-Net are initialized to zero. We train~\methodname{} on~\datasetname{}, using the original input image resolution of $512 \times 512$. We utilize the AdamW optimizer with a learning rate $1e-5$. We train our model for $20,000$ steps on $8$ NVIDIA A6000 GPUs with an effective batch size of $16$, which takes around $12$ hours. During training, we randomly drop text prompts $20\%$ of the time to allow the model to take cues from the input depth map. We find the checkpoint at $15,000$ to produce the best qualitative results and use it for further inference.
We also run an additional experiment where we make the generation U-Net trainable. We call this model\textbf{~\methodname{}$^{\text{*}}$}. We use the same training hyper-parameters and consider the checkpoint at $17,000$ steps. 
From Fig.~\ref{fig:supp_cmp_set_1} and Fig.~\ref{fig:supp_cmp_set_2}, we can see improved results compared to the frozen generation U-Net. However, the VRAM requirements and training time almost double, due to the increase in the number of trainable parameters.

\subsection{Training Details for Baseline Methods}
\label{subsec_appendix:baselines_training}
\noindent
\textbf{Fine-tuning of BrushNet~\cite{ju2024brushnet}.}
Keeping the generation U-Net frozen, we fine-tune BrushNet using the input mask and masked image using the same hyperparameters used to train~\methodname{}. We do not randomly drop text prompts and select the checkpoint at $17,000$ steps for evaluation. We refer to this model as \textbf{\textit{``BrushNet-FT''}} in Sec.~\ref{sec:exp_results} of the main paper and compare our results against it. We found that initializing the weights from the Stable Diffusion v1.5~\cite{rombach2022high} checkpoint was superior as compared to initializing from the pre-trained BrushNet~\cite{ju2024brushnet} checkpoint.

\subsection{Inference Details}
\label{subsec_appendix:inference_details}
During inference, we set the classifier free guidance scale (CFG) to $7.5$ and use the UniPC scheduler~\cite{zhao2024unipc} for $50$ time-steps across all experiments.

\clearpage

\section{Additional Results}

\subsection{More Results on Google Scanned Objects (GSO)}
\label{subsec_appendix:supp_gso}
\begin{figure}[!t]
    \centering
    \includegraphics[width=\linewidth]{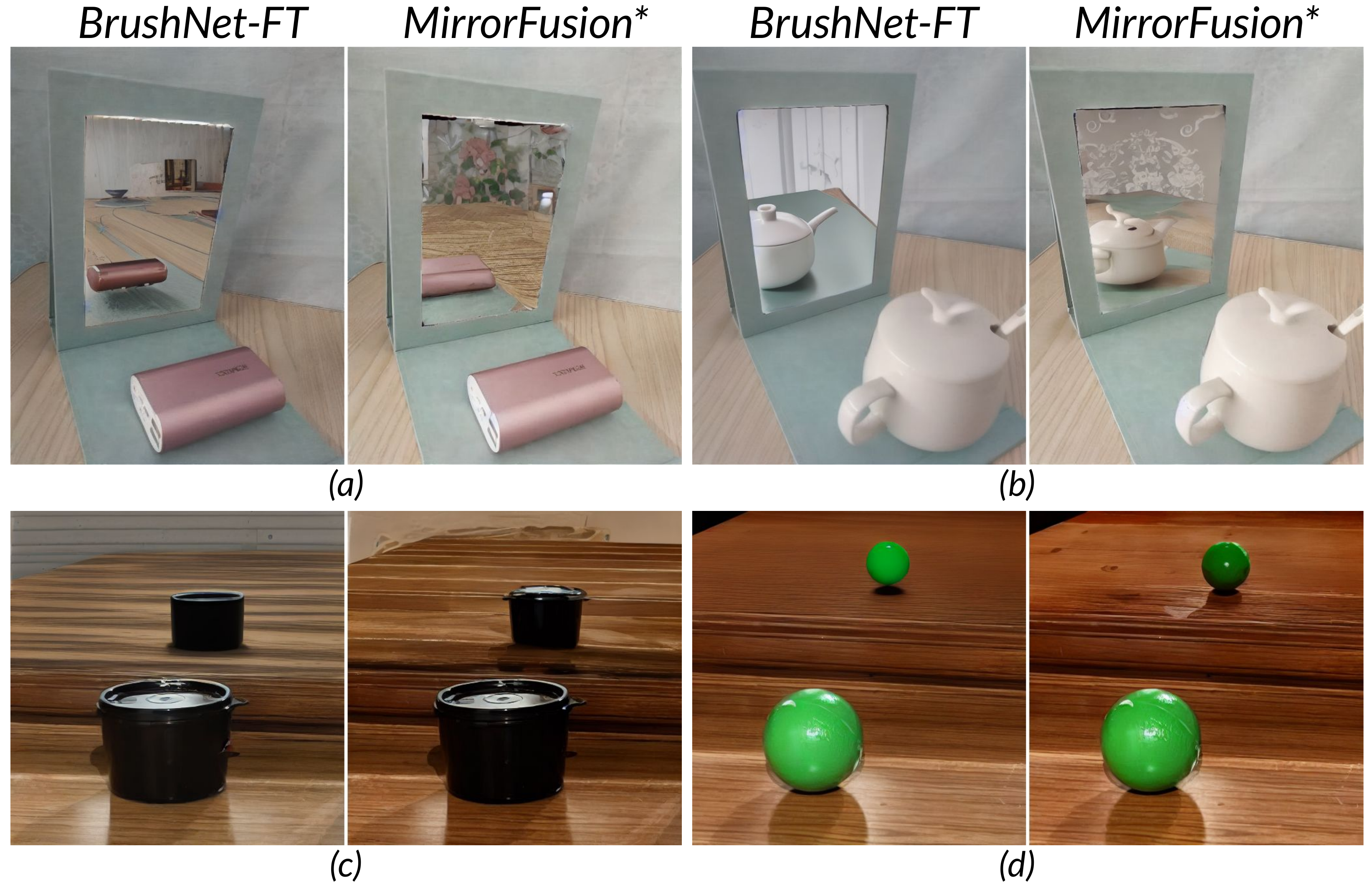}
    \caption{\textbf{Performance on Real-world scenes} We show results on images from MSD~\cite{Yang_2019_ICCV} dataset (a)~\& (b) and examples from images captured using a smartphone device (c)~\& (d). Appendix~\ref{subsec_appendix:real_world} describes the experimental details and text prompts used for the inference. We observe that ``BrushNet-FT'' does not generate accurate reflections, whereas our method is able to generate plausible reflections on the mirror.}
    \label{fig:supp_real_world}
\end{figure}
We provide additional results on 3D models from Google Scanned Objects (GSO)~\cite{downs2022google} in Fig.~\ref{fig:supp_gso}. GSO contains real-world scanned objects. We create renderings using these objects with the pipeline discussed in Sec.~\ref{sec:dataset}. We notice that our method ~\methodname{}$^{\text{*}}$
consistently generates accurate reflections of objects and floors in the mirror. However,  \textit{``BrushNet-FT''}, is not able to generate the reflection of the floor correctly in image with blue ball (Fig.~\ref{fig:supp_gso} (o), and (p)) and carton (Fig.~\ref{fig:supp_gso} (l))
Further, it does not get the appearance of the pencil-box right, as shown in Fig.~\ref{fig:supp_gso} (g) and (h). Additionally, it generates the reflection with the wrong structure in Fig.~\ref{fig:supp_gso} (c) and (d). These results further substantiate the generalization capabilities of our method.

Text prompts used for results in Fig.~\ref{fig:supp_gso} are as follows:
\begin{itemize}
    \item \textbf{(a)~\& (b).} \textit{``A perfect plane mirror reflection of a sofa with purple cushioning.''}
    \item \textbf{(c)~\& (d).} \textit{``A perfect plane mirror reflection of a yellow chair.''}
    \item \textbf{(e)~\& (f).} \textit{``A perfect plane mirror reflection of a white stool with a purple top.''}
    \item \textbf{(g)~\& (h).} \textit{``A perfect plane mirror reflection of a purple bag with bluish circular patterns.''}
    \item \textbf{(i)~\& (j).} \textit{``A perfect plane mirror reflection of a camouflaged military-style bag.''}
    \item \textbf{(k)~\& (l).} \textit{``A perfect plane mirror reflection of a cardboard box on a patterned floor.''}
     \item \textbf{(m)~\& (n).} \textit{``A perfect plane mirror reflection of a yellow and white mug on a grey surface.''}
      \item \textbf{(o)~\& (p).} \textit{``A perfect plane mirror reflection of a blue ball with an orange cover.''}
\end{itemize}

\begin{figure}[!t]
    \centering  \includegraphics[width=\linewidth]{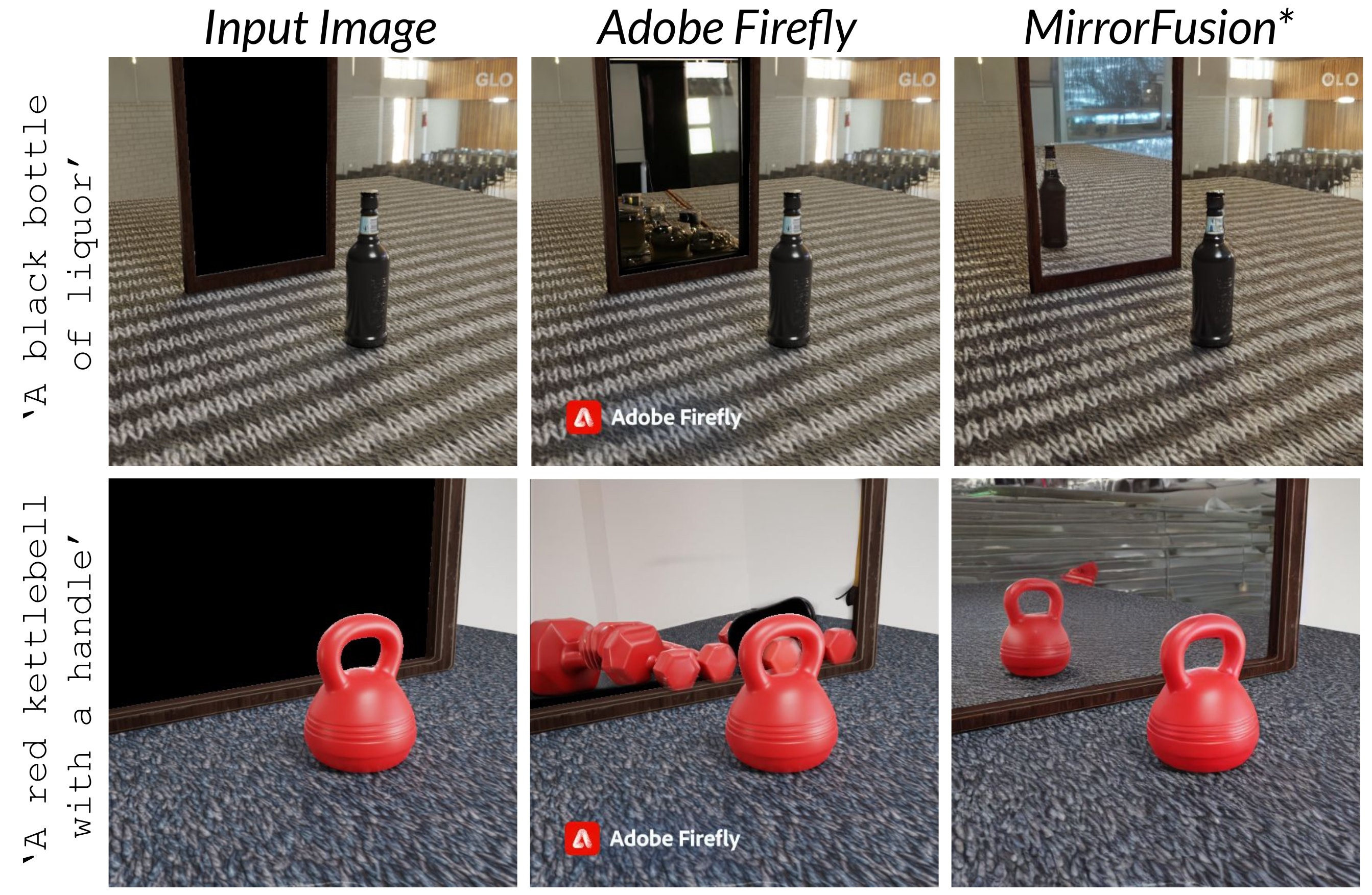}
    \caption{\textbf{Qualitative Comparison with Commercial Products} We compare our results with Adobe Firefly. Our method is significantly better than the existing commercial product. This highlights the challenging nature of the task and the effectiveness of our proposed method in addressing it.}
    \label{fig:supp_commercial}
    \vspace{-2em}
\end{figure}

\subsection{Results on real-world scenes.}
\label{subsec_appendix:real_world}
We present real-world examples from the MSD~\cite{Yang_2019_ICCV} dataset in Fig.~\ref{fig:supp_real_world} (a) and (b), utilizing the ground truth (GT) masks provided within the dataset as the corresponding mirror masks. Since our method requires depth, we infer it from Marigold and normalize it as described in Sec.~\ref{subsec:depth_mask_guidance}. We observe that the baseline method fails to position the object accurately and produces incorrect color in Fig.~\ref{fig:supp_real_world} (a). In contrast, our method generates better reflections on the mirror.

We also capture more examples from a hand-held smartphone device in Fig.~\ref{fig:supp_real_world} (c) \& (d). We manually annotate the mask corresponding to the mirror location and infer the depth from Marigold~\cite{ke2024repurposing} as described above. Similar to the previous observation, our method preserves the shape of the object. Check the lid in Fig.~\ref{fig:supp_real_world} (c) and the roundness of the ball in Fig.~\ref{fig:supp_real_world} (d). These results show that our method generates better reflections than the baselines on real-world settings. Our method shows promising results on real-world settings, but still has scope for improvement, showing the challenging nature of this task.

Text prompts used for generating the results in Fig.~\ref{fig:supp_real_world} are as follows:
\begin{itemize}
    \item \textbf{(a).} \textit{``A perfect plane mirror reflection of a rose gold colored portable power-bank.''}
    \item \textbf{(b).} \textit{``A perfect plane mirror reflection of a white ceramic teapot.''}
    \item \textbf{(c).} \textit{``A perfect plane mirror reflection of a black round box with a black lid on it.''}
    \item \textbf{(d).} \textit{``A perfect plane mirror reflection of a green color round ball.''}
\end{itemize}

\begin{figure*}[!t]
    \centering
    \includegraphics[width=\linewidth]{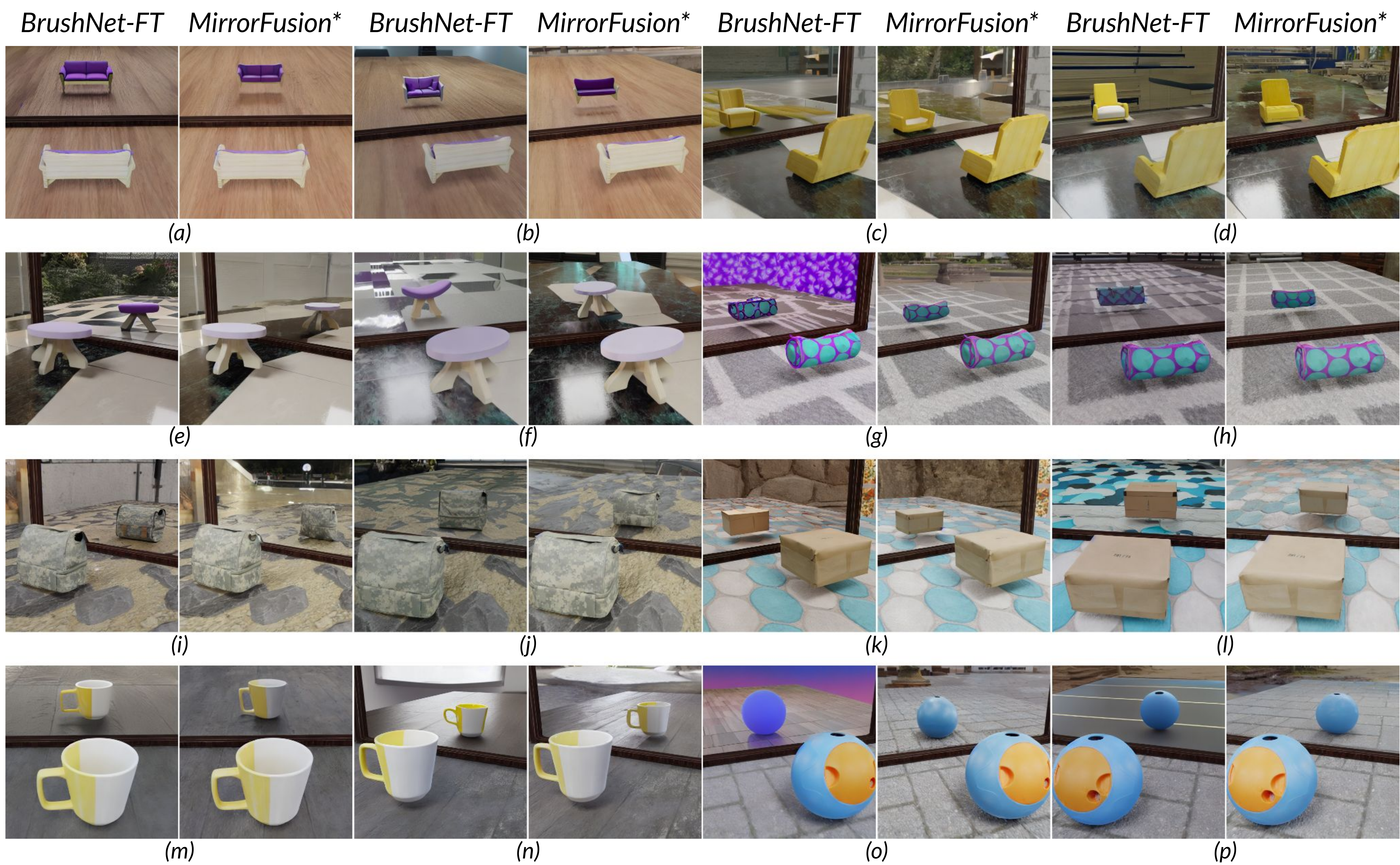}
    \caption{\textbf{Qualitative Comparison on unseen 3D assets from GSO.} We show results from (a)~\& (b) ``3D Dollhouse Sofa'', (c)~\& (d) ``3D Dollhouse Swing'', (e)~\& (f) ``3D Dollhouse TablePurple'', (g)~\& (h) ``Big Dot Aqua Pencil Case'', (i)~\& (j) ``Digital Camo Double Decker Lunch Bag'', (k)~\& (l) ``INTERNATIONAL PAPER Willamette 4 Brown Bag'' , (m)~\& (n) ``Room Essentials Mug White Yellow'' and (o)~\& (p) ``Toys R Us Treat Dispenser Smart Puzzle Foobler''. Appendix~\ref{subsec_appendix:supp_gso} describes how images are generated and text-prompts used for the inference. We observe that ``BrushNet-FT'' does not generate accurate reflections in (c),(d),(f),(g),(h) whereas our method is able to generate correct reflections on the mirror. }
    \label{fig:supp_gso}
\end{figure*}

\FloatBarrier

\subsection{Comparison with Commercial Products.}
\label{subsec_appendix:cmp_firefly}
We compare our method with commercial products such as Adobe Firefly in Fig.~\ref{fig:supp_commercial}. Our method significantly outperforms existing commercial solutions. Results from Fig.~\ref{fig:supp_commercial} highlight the challenging nature of the task of generating plausible mirror reflections and the critical gap that exists in current state-of-the-art methods. Text prompts used in Fig.~\ref{fig:supp_commercial} are as follows:
\begin{itemize}
    \item \textbf{$1^{st}$ row.} \textit{``A perfect plane mirror reflection of a black bottle of liquor.''}
    \item \textbf{$2^{nd}$ row.} \textit{``A perfect plane mirror reflection of a red kettle-ball with a handle.''}
\end{itemize}

\subsection{Robustness to pre-trained monocular depth estimation methods}
\label{subsec_appendix:depth_algo_choice}
\begin{figure}[!t]
    \centering
    \includegraphics[width=\linewidth]{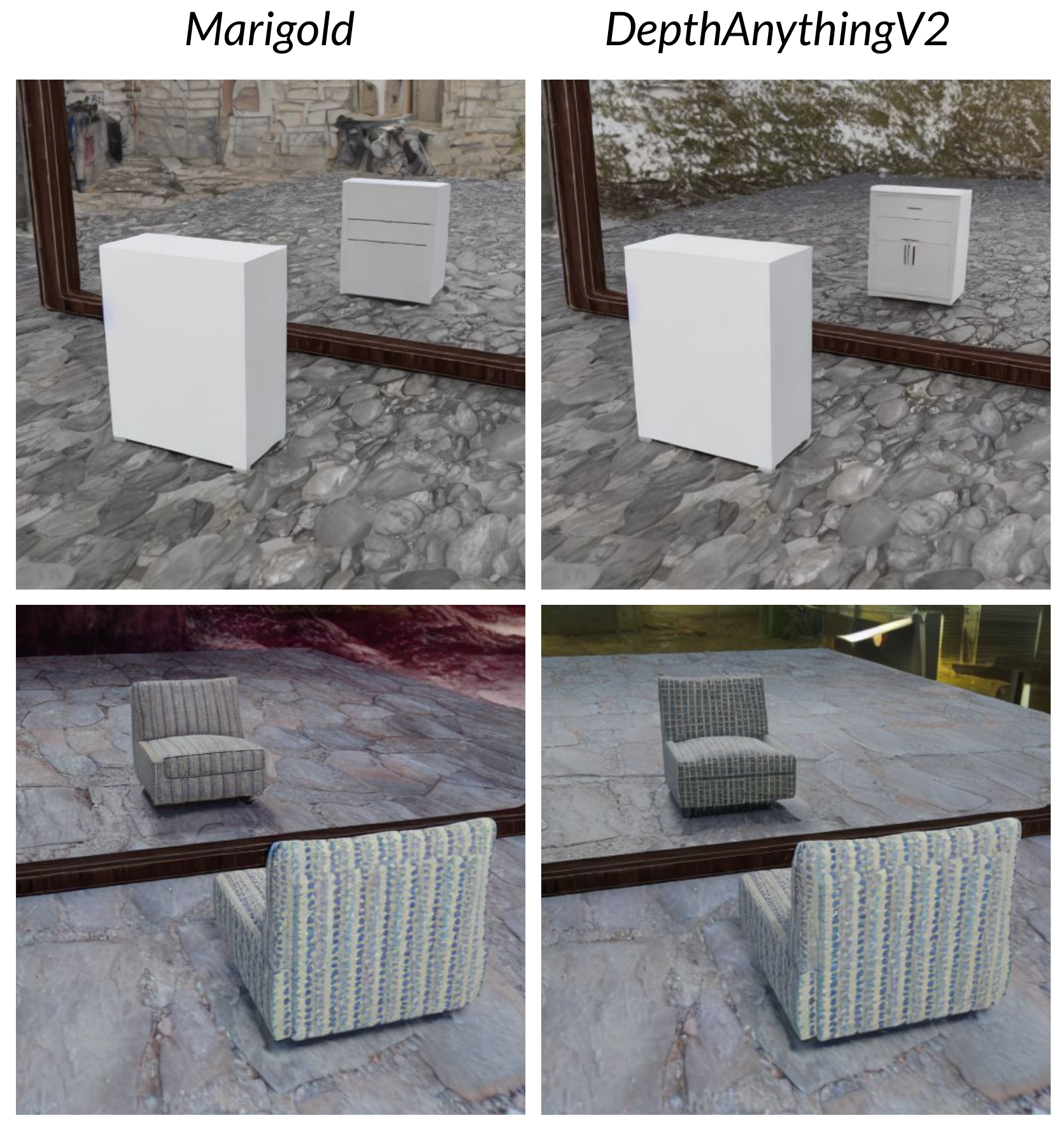}
    \caption{\textbf{Choice of pre-trained monocular depth estimation method during inference.} We observe negligible differences in the reflection generation across both choices, Marigold~\cite{ke2024repurposing} and DepthAnythingV2~\cite{yang2024depth}, supporting the stability of our method regardless of the chosen pre-trained monocular depth estimation technique. We use ``Marigold'' in all our experiments.}
    \label{fig:supp_pretrain_choice}
\end{figure}

Our method is invariant to the choice of the pre-trained monocular depth estimation method. We present results from two state-of-the-art methods, Marigold~\cite{ke2024repurposing} and DepthAnythingV2~\cite{yang2024depth}, in Fig.~\ref{fig:supp_pretrain_choice}. We observe minimal variation in the generation of reflections between both options, thereby confirming the robustness of our approach to the preference of the pre-trained monocular depth estimation method. 

Text prompts for Fig.~\ref{fig:supp_pretrain_choice} are as follows, each row uses the same text prompt:
\begin{itemize}
    \item \textbf{$1^{st}$ row.} \textit{``A perfect plane mirror reflection of a rectangular cabinet with a door, two drawers, a truncated triangular base, and a triangular top.''}
    \item \textbf{$2^{nd}$ row.} \textit{``A perfect plane mirror reflection of a swivel chair  with curved backrest, slanted seat, curved armrests, and a triangular top.''}
\end{itemize}

\subsection{More Qualitative Comparisons}
\begin{figure*}[!t]
    \centering
    \includegraphics[width=\linewidth]{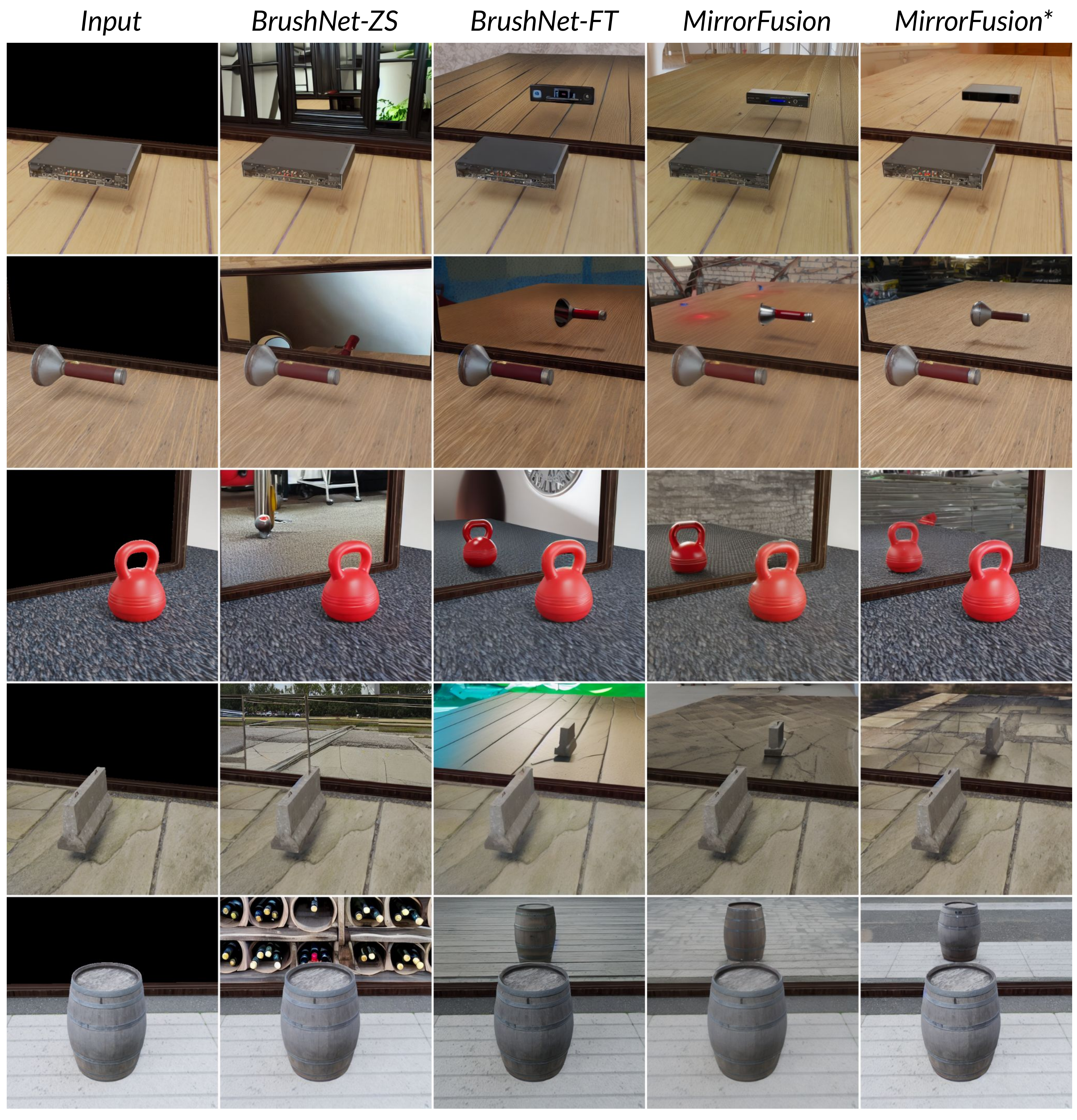}
    \caption{\textbf{Qualitative Comparison.} We observe that the state-of-the-art inpainting method ``BrushNet-ZS'' is not able to generate plausible reflections \textbf{($2^{nd}$ column)}. ``BrushNet-FT'' which is fine-tuned on~\datasetname{} is able to generate plausible reflections, \textbf{$3^{rd }$ column}, but fails to accurately get the shape of the object. For example, the top surface of ``dvd-player'' in $1^{st}$ row is completely missing. The ''flashlight'' reflection’s structure and appearance do not correspond with the object ($2^{nd}$ row). Compared to these baselines~\methodname{} generates plausible reflections. Still there is issue in the shape of the ``flashlight'' in $2^{nd}$ row. These issues are mitigated by ~\methodname{}$^{\text{*}}$, which generates realistic, plausible and geometrically accurate reflections on the mirror.   }
    \label{fig:supp_cmp_set_1}
\end{figure*}

\begin{figure*}[!t]
    \centering
    \includegraphics[width=\linewidth]{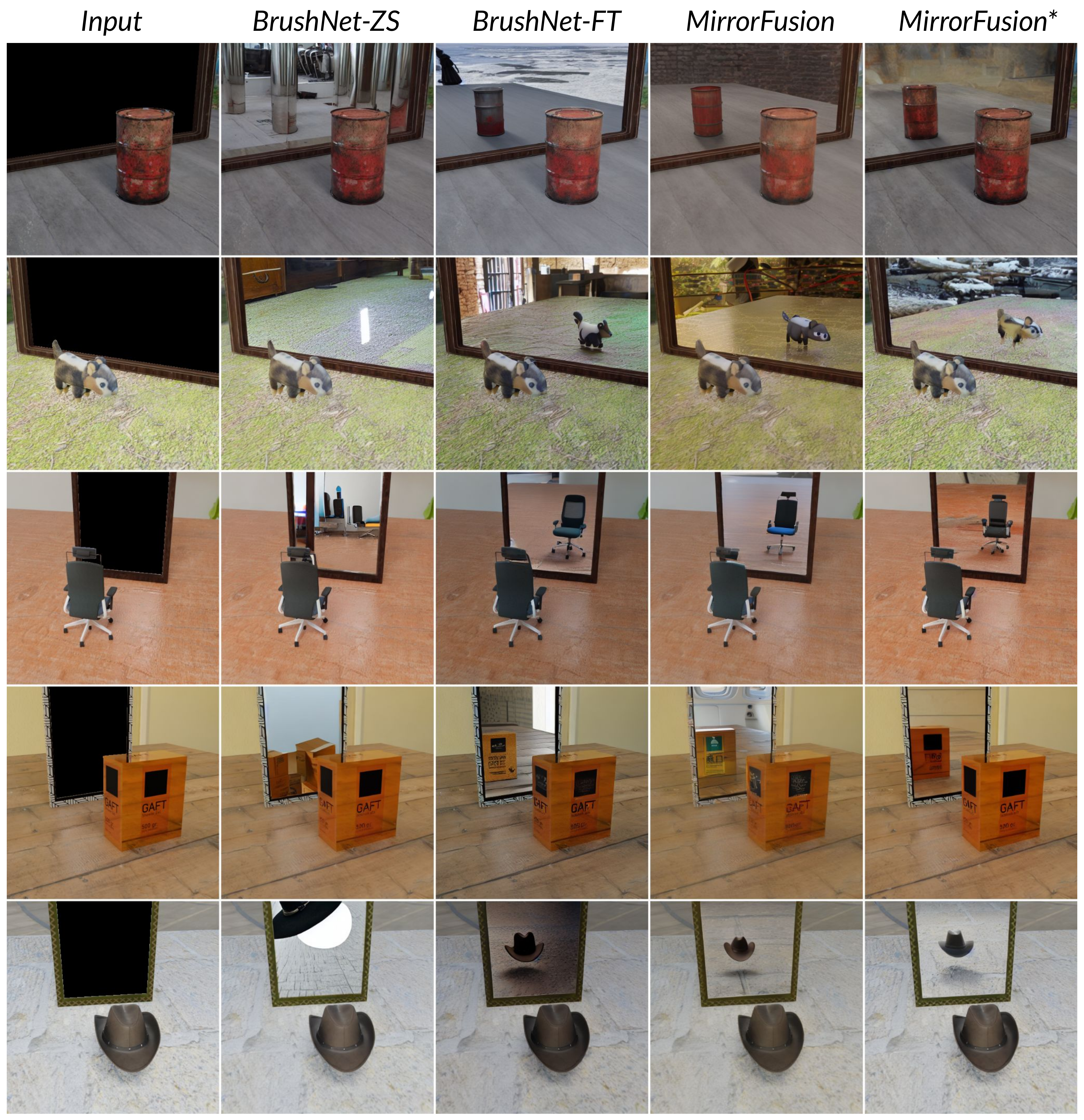}
    \caption{\textbf{Qualitative Comparison.} Similar to the observation in Fig.~\ref{fig:supp_cmp_set_1}, we observe that the state-of-the-art inpainting method ``BrushNet-ZS'' is not able to generate plausible reflections \textbf{($2^{nd}$ column)}. ``BrushNet-FT'' which is fine-tuned on~\datasetname{} is able to generate plausible reflections, \textbf{$3^{rd }$ column}  but fails to get shape of the object in the reflection. For example, observe the ``chair'' in $3^{rd}$ row, the head of the chair is missing. The pose of the toy in $2^{nd}$ row does not correspond to that of the real object. Compared to this~\methodname{} and ~\methodname{}$^{\text{*}}$ generates plausible reflections on the mirror. }
    \label{fig:supp_cmp_set_2}
\end{figure*}

As discussed in Sec.~\ref{sec:exp_results}, we compare our method with zero-shot baselines, denoted by \textit{``-ZS''} and baselines trained on~\datasetname{}, denoted by \textit{``-FT''}. We provide additional results in Fig.~\ref{fig:supp_cmp_set_1} and ~\ref{fig:supp_cmp_set_2}. Consistent with the findings in the main paper, our method generates better mirror reflections while preserving the fidelity of both the object's appearance and the floor.


\textbf{Fig.~\ref{fig:supp_cmp_set_1}} Each row in this figure uses the same text prompt. Text prompts are as follows:
\begin{itemize}
    \item \textbf{$1^{st}$ row.} \textit{``A perfect plane mirror reflection of a multifunctional electronic device, including HDMI Blu-ray player, stereo receiver, amplifier, CD, and DVD player.''}
    \item \textbf{$2^{nd}$ row.} \textit{``A perfect plane mirror reflection of a red flashlight with a metal pipe.''}
    \item \textbf{$3^{rd}$ row.} \textit{``A perfect plane mirror reflection of a red kettlebell with a handle.''}
    \item \textbf{$4^{th}$ row.} \textit{``A perfect plane mirror reflection of a concrete block.''}
    \item \textbf{$5^{th}$ row.} \textit{``A perfect plane mirror reflection of a wooden barrel.''}
\end{itemize}

\textbf{Fig.~\ref{fig:supp_cmp_set_2}} Each row in this figure uses the same text prompt. Text prompts are as follows:
\begin{itemize}
    \item \textbf{$1^{st}$ row.} \textit{``A perfect plane mirror reflection of a large, red, rusty metal barrel.''}
    \item \textbf{$2^{nd}$ row.} \textit{``A perfect plane mirror reflection of a small stuffed animal toy.''}
    \item \textbf{$3^{rd}$ row.} \textit{``A perfect plane mirror reflection of a modern office chair with a blue upholstered seat, back, and headrest.''}
    \item \textbf{$4^{th}$ row.} \textit{``A perfect plane mirror reflection of a Gaft Shower Gel Box.''}
    \item \textbf{$5^{th}$ row.} \textit{``A perfect plane mirror reflection of a black cowboy hat.''}
\end{itemize}

\section{Limitations and Social Impact}
\label{appendix:limitations}
\textbf{Limitations.} As our method leverages synthetic data to train a model capable of producing realistic mirror reflections, the model still has scope for improvement in generating reflections for highly complex objects and scenarios. Although our model generates plausible results on real-world images, there is significant scope for improvement, which can be achieved by using more advanced photo-realistic simulators or collecting large-scale real-world images. We aim to address these issues in our future work.

\vspace{2mm} 
\noindent \textbf{Social Impact.} Our method uses diffusion-based generative models, which, despite their potential, can be exploited for spreading misinformation. Therefore, it is crucial to use these models responsibly.

\section{Additional Details}
\label{appendix:add_details}

\subsection{Results from recent T2I methods}
\label{subsec_appendix:failure_cases_t2i}
\begin{figure}[!t]
    \centering
    \includegraphics[width=\linewidth]{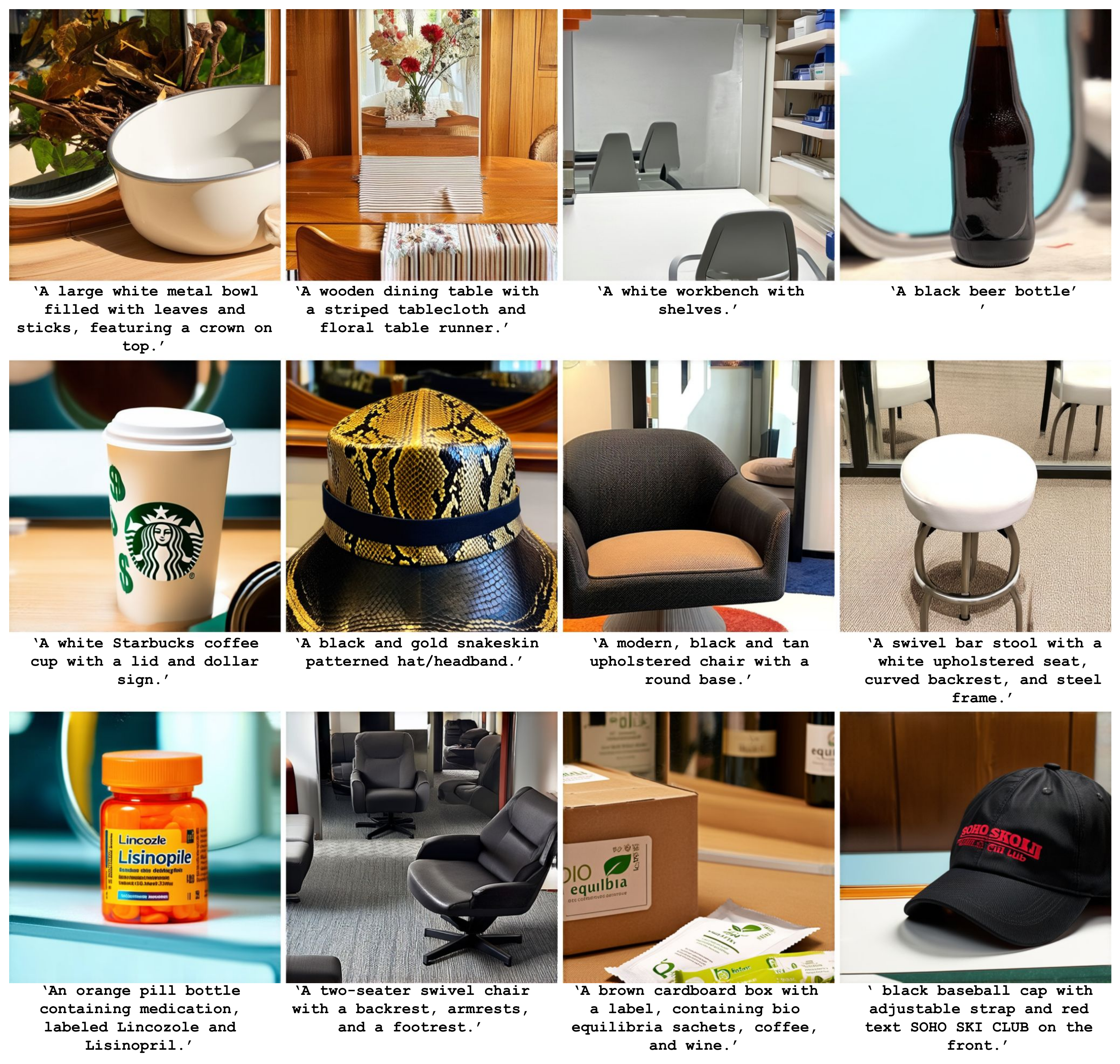}
    \caption{\textbf{Additional results of images generated from Stable Diffusion 3~\cite{SD3}.} Text-to-image models struggle to produce consistent and controlled mirror reflections when prompted to generate them. We use the prefix \textit{``A perfect plane mirror reflection of''} and suffix \textit{``in front of the mirror positioned at an angle with respect to the mirror.''} along with the object description.}
    \label{fig:supp_t2i_results}
\end{figure}

We present additional results from the recent Stable Diffusion 3~\cite{SD3} model in Fig.~\ref{fig:supp_t2i_results}.
Text prompts are generated by using the prefix: \textit{``A perfect plane mirror reflection of''} and suffix: \textit{``in front of the mirror positioned at an angle with respect to the mirror.''} to the object description of the input image. We observe that standalone text-to-image methods are inadequate in generating controlled and realistic mirror reflections.

\subsection{Text prompts used in the experiments}   
\label{subsec_appendix:captions_used_in_exp}
This section provides the text prompts for the image generations in the main paper.

\noindent
\textbf{Figure 1.} Each row in this figure uses the same text prompt. Text prompts are as follows:
\begin{itemize}
    \item \textbf{First row.} \textit{``A perfect plane mirror reflection of a swivel chair with a curved backrest, slanted seat, slender metal frame, and padded seat and backrest.''}
    \item \textbf{Second row.} \textit{``A perfect plane mirror reflection of a large red, yellow, and black industrial cement mixer.''}
\end{itemize}

\noindent
\textbf{Figure 2.} Text prompts are already mentioned in the Figure of the main paper.

\noindent
\textbf{Figure 5.} Text prompts are as follows:
\begin{itemize}
    \item \textbf{(a).} \textit{``A perfect plane mirror reflection of a white golf ball with a red stripe and the letter O on it.''}
    \item \textbf{(b).} \textit{``A perfect plane mirror reflection of a chair with a curved slatted frame, tufted backrest, and curved seat.''}
\end{itemize}

\noindent
\textbf{Figure 6.} Text prompts are as follows:
\begin{itemize}
    \item \textbf{(a).} \textit{``A perfect plane mirror reflection of a modern wooden chaise lounge with a white cushion.''}
    \item \textbf{(b).} \textit{``A perfect plane mirror reflection of a swivel chair with a curved backrest, slender armrest, and swivel base.''}
    \item \textbf{(c).} \textit{``A perfect plane mirror reflection of a black cylindrical with a lid.''}
    \item \textbf{(d).} \textit{``A perfect plane mirror reflection of a wooden box with intricate floral and heart-shaped carvings on each side, featuring a dark brown hue with visible wood grain texture.''}
\end{itemize}

\textbf{Figure 7.} Each row in this figure uses the same text prompt. Text prompts are as follows:
\begin{itemize}
    \item \textbf{$1^{st}$ row.} \textit{``A perfect plane mirror reflection of a large red, yellow, and black industrial cement mixer.''}
    \item \textbf{$2^{nd}$ row.} \textit{``A perfect plane mirror reflection of a gold lipstick container.''}
    \item \textbf{$3^{rd}$ row.} \textit{``A perfect plane mirror reflection of a cylindrical object with a cream-colored exterior and a central hollow core; vertical seams divide the outer surface.''}
    \item \textbf{$4^{th}$ row.} \textit{``A perfect plane mirror reflection of a weathered wooden treasure chest with metal reinforcements, large metal ring on the side, and mossy accents.''}
    \item \textbf{$5^{th}$ row.} \textit{``A perfect plane mirror reflection of a grey cabinet with gold legs and chest of drawers.''}
    \item \textbf{$6^{th}$ row.} \textit{``A perfect plane mirror reflection of a black stone with intricate swirl designs on it.''}
\end{itemize}

\textbf{Figure 8.} Each row in this figure uses the same text prompt. Text prompts are as follows:
\begin{itemize}
    \item \textbf{$1^{st}$ row.} \textit{``A perfect plane mirror reflection of a slanted-top cuboid footstool.''}
    \item \textbf{$2^{nd}$ row.} \textit{``A perfect plane mirror reflection of a footstool with a cuboid base, spherical top, seat, and backrest.''}
\end{itemize}

\textbf{Figure 9.}Text prompts are as follows:
\begin{itemize}
    \item \textbf{(a).} \textit{``A perfect plane mirror reflection of a white ceramic bowl on a textured gray surface..''}
    \item \textbf{(b).} \textit{``A perfect plane mirror reflection of a camouflaged military-style bag''}
\end{itemize}
\subsection{Generation of Segmentation Masks for computing metrics}
\label{subsec_appendix:seg_masks}
We compare the accuracy of the geometry of the generated reflection by comparing IoU between the segmentation mask of the reflection in the ground-truth object and the segmentation mask of the reflection in the generated object in Sec.~\ref{sec:exp_results}. We utilize SAM to generate these segmentation masks. We provide initial seed points to SAM~\cite{kirillov2023segment} along with a rough bounding box. SAM then segments out the reflection of the object in ground truth as well as the generated image. Camera viewpoint variations within our dataset pose a challenge for reliable seed point initialization. We address this by manually creating a mapping to select seed points based on the camera pose. To accelerate the evaluation, we cache the segmentation masks of the ground-truth images.

\onecolumn
\twocolumn

\end{document}